\documentclass[letterpaper]{article} 
\usepackage{aaai2026}  
\usepackage{times}  
\usepackage{helvet}  
\usepackage{courier}  
\usepackage[hyphens]{url}  
\usepackage{graphicx} 
\usepackage{natbib}  
\usepackage{caption} 
\usepackage{algorithm}
\usepackage{algorithmic}

\usepackage{newfloat}
\usepackage{listings}
\DeclareCaptionStyle{ruled}{labelfont=normalfont,labelsep=colon,strut=off} 
\floatstyle{ruled}
\newfloat{listing}{tb}{lst}{}
\floatname{listing}{Listing}
\usepackage[table,xcdraw]{xcolor}
\usepackage[utf8]{inputenc} 
\usepackage{booktabs}       
\usepackage{amsfonts}       
\usepackage{nicefrac}       
\usepackage{microtype}      
\usepackage{xspace}
\usepackage{tablefootnote}
\usepackage{mdframed}
\usepackage{placeins}
\usepackage{multirow}
\usepackage{mathtools}
\usepackage{booktabs}

\usepackage[most]{tcolorbox}

\usepackage{lipsum}
\usepackage{etoolbox}
\AtBeginEnvironment{tcolorbox}{\fontsize{9}{10}\selectfont}

\usepackage{tabularx}

\usepackage{amsthm}
\newtheorem{definition}{Definition}[section]

\usepackage{tikz}
\usepackage{dsfont} 

\usepackage{pifont}
\usepackage{mathabx}

\newcommand{\modelname}{StepFun-Formalizer\xspace}
\newcommand{\xname}{ThinkingF\xspace}

\newcommand{\NM}{NuminaMath-1.5\xspace}
\newcommand{\KMAuto}{Kimina-Autoformalizer\xspace}
\newcommand{\Claude}{Claude 3.7 Sonnet\xspace}
\newcommand{\RR}{DeepSeek-R1\xspace}
\newcommand{\RRD}{DeepSeek-R1-Distill-Qwen\xspace}

\newcommand{\FMLite}{FormalMATH-Lite\xspace}
\newcommand{\PBench}{ProverBench\xspace}
\newcommand{\CBench}{CombiBench\xspace}
\newcommand{\indicator}[1]{\mathds{1}{#1}}
\newcommand{\GR}{\textasciigrave}
\definecolor{codegreen}{rgb}{0,0.6,0}
\definecolor{codegray}{rgb}{0.5,0.5,0.5}
\definecolor{codepurple}{rgb}{0.58,0,0.82}
\definecolor{backcolour}{rgb}{0.91,0.91,0.9}
\definecolor{shallowRed}{rgb}{1, 0.8, 0.8}
\definecolor{shallowYellow}{rgb}{1, 0.953, 0.8}
\definecolor{shallowBlue}{rgb}{0.8, 0.8, 1}
\definecolor{orange}{rgb}{1, 0.6, 0}
\definecolor{shallowOrange}{rgb}{1, 0.88, 0.7}

\definecolor{keywordcolor}{rgb}{0.7, 0.1, 0.1}   
\definecolor{commentcolor}{rgb}{0.4, 0.4, 0.4}   
\definecolor{symbolcolor}{rgb}{0.0, 0.1, 0.6}    
\definecolor{sortcolor}{rgb}{0.1, 0.5, 0.1}      
\definecolor{errorcolor}{rgb}{1, 0, 0}           
\definecolor{stringcolor}{rgb}{0.5, 0.3, 0.2}    

\definecolor{leftcolor}{rgb}{0.522, 0.765, 0.863}  
\definecolor{midcolor}{rgb}{0.855, 0.776, 0.812}
\definecolor{rightcolor}{rgb}{0.886, 0.753, 0.596}

\newcommand{\circled}[1]{\tikz[baseline=(char.base)]{
    \node[shape=circle, fill=black, inner sep=0.8pt] (char) {\color{white}\scriptsize #1};
}}

\newcommand{\UCAS}{University of Chinese Academy of Sciences}
\newcommand{\SKLP}{SKL of Processors, Institute of Computing Technology, CAS}

\newcommand{\USTC}{University of Science and Technology of China}
\newcommand{\STEPFUN}{StepFun Inc.}

\title{\modelname: Unlocking the Autoformalization Potential \\ of LLMs Through Knowledge-Reasoning Fusion}
\author{
    Yutong~Wu\textsuperscript{\rm 1, \rm 2}\equalcontrib,
    Di~Huang\textsuperscript{\rm 1},
    Ruosi~Wan\textsuperscript{\rm 4}, 
    Yue~Peng\textsuperscript{\rm 4},
    Shijie~Shang\textsuperscript{\rm 4} \\
    Chenrui~Cao\textsuperscript{\rm 1, \rm 3},
    Lei~Qi\textsuperscript{\rm 1, \rm 3},
    Rui~Zhang\textsuperscript{\rm 1},
    Zidong~Du\textsuperscript{\rm 1},
    Jie~Yan\textsuperscript{\rm 4},
    Xing~Hu\textsuperscript{\rm 1}\thanks{Corresponding author} \\
}
\affiliations{
    \textsuperscript{1}\SKLP{}  \\
    \textsuperscript{2}\UCAS{} \quad \textsuperscript{3}\USTC{} \\ 
    \textsuperscript{4}\STEPFUN{} \\
    Contact: wuyutong22s@ict.ac.cn
}

\usepackage{bibentry}
\usepackage{booktabs} 
\usepackage{caption}  
\usepackage{makecell}

\begin{document}

\maketitle

\begin{abstract}

Autoformalization aims to translate natural-language mathematical statements into a formal language. While LLMs have accelerated progress in this area, existing methods still suffer from low accuracy. We identify two key abilities for effective autoformalization: comprehensive mastery of formal-language domain knowledge, and reasoning capability of natural language problem understanding and informal-formal alignment. Without the former, a model cannot identify the correct formal objects; without the latter, it struggles to interpret real-world contexts and map them precisely into formal expressions.
To address these gaps, we introduce \xname, a data synthesis and training pipeline that improves both abilities. First, we construct two datasets: one by distilling and selecting large-scale examples rich in formal knowledge, and another by generating informal-to-formal reasoning trajectories guided by expert-designed templates. We then apply SFT and RLVR with these datasets to further fuse and refine the two abilities. The resulting 7B and 32B models exhibit both comprehensive formal knowledge and strong informal-to-formal reasoning. Notably, \modelname-32B achieves SOTA BEq@1 scores of 40.5\% on \FMLite and 26.7\% on \PBench, surpassing all prior general-purpose and specialized models.
\end{abstract}

\begin{links}
    \link{Code}{https://github.com/stepfun-ai/StepFun-Formalizer}
    \link{Models}{https://huggingface.co/stepfun-ai/StepFun-Formalizer-32B}
    \link{Extended version}{https://arxiv.org/abs/2508.04440}
\end{links}

\section{Introduction}
\label{sec:introduction}

Autoformalization aims to translate natural-language mathematical statements into formally verifiable statements in formal languages such as Lean \cite{Lean}, Coq \cite{coq}, and Isabelle \cite{isabelle}. With recent advances in automated theorem proving \cite{ATP_survey} and formal verification \cite{FM_survey}, it has garnered growing interest \cite{autoformalization_survey} for underpinning data synthesis for theorem provers \cite{Prover-DS-V1,Prover-DS-V1.5,Prover-GD,Prover-Leanabell-V1,Prover-Kimina,Prover-DS-V2,Prover-Leanabell-V2}, the validation of informal mathematical reasoning \cite{donttrustverify}, and the generation of verifiable code \cite{fvel,clever,veribench}.


The current mainstream approaches for autoformalization involve employing an LLM to translate informal mathematical problems into their corresponding formal statements. These approaches can be categorized into two types: (1) Directly distilling or prompting general-purpose LLMs, such as FormalMATH \cite{FormalMath}, which generates autoformalization training data using GPT-4, and FMC \cite{FMC}, which directly employs \RR \cite{deepseekr1} as the translation model. (2) Training a specialized model for the autoformalization task from scratch, starting with existing manually annotated informal-formal pairs and followed by supervision from human experts, such as Lean Workbook~\cite{LeanWorkbook} and \KMAuto~\cite{Prover-Kimina}. However, these two types of methods suffer from low accuracy and require substantial human effort to check and revise the human-generated formal statements \cite{FormalMath, formalizingcomplex}.

\begin{figure*}[t]
    \centering

    \begin{tcolorbox}[
        colback=white,      
        colframe=white,     
        arc=0pt,            
        boxrule=1pt,        
        width=\linewidth,
        center,             
        left=0mm,           
        right=0mm,          
        top=0mm,            
        bottom=0mm,         
    ]
    \setlength{\fboxsep}{0pt}
    \selectfont
    \textbf{Informal Problem}: Victoria wants to order at least 550 donuts from Dunkin' Donuts for the HMMT 2014 November contest. However, \setlength{\fboxsep}{1pt}\fbox{donuts only come in multiples of twelve}. Assuming every twelve donuts cost \$7.49, what is the minimum amount Victoria needs to pay, in dollars? Prove that the answer is: 344.54.
    
    \noindent
\begin{minipage}[t]{0.52\textwidth}
    \vspace{0pt}
    
    \begin{tcolorbox}[
        colback=white,      
        colframe=black,     
        arc=5pt,            
        boxrule=1pt,        
        width=\linewidth,   
        center,             
        left=0mm,           
        right=0mm,          
        top=0mm,            
        bottom=0mm,         
    ]
    \selectfont
    \textbf{Claude4-thinking} {(advanced general-purpose models, lack of formal knowledge)}:
    
    \textless think\textgreater\\
Let me think about this problem step by step.
Victoria wants to order at least 550 donuts. Donuts come in multiples of 12...

    \textless /think\textgreater
    \begin{lstlisting}[language=lean,basicstyle=\ttfamily,frame=none,escapechar=!,breakindent=0pt]
theorem b : 
  let target_donuts : ℕ := 550
  let donuts_per_dozen : ℕ := 12
  let min_dozens := Nat.ceil_div target_donuts donuts_per_dozen
  ...
    \end{lstlisting}

    \vspace{-30pt}
    \hspace{238pt}\selectfont\textcolor[RGB]{0, 0, 0}{\fontsize{25}{0}\ding{55}}
    
    \textbf{Mistakes}: Function \texttt{Nat.ceil\_div} does not exist in Lean 4. 
    
    \end{tcolorbox}

    \vspace{-7pt}

    \begin{tcolorbox}[
        colback=white,      
        colframe=black,     
        arc=5pt,            
        boxrule=1pt,        
        width=\linewidth,   
        center,             
        left=0mm,           
        right=0mm,          
        top=0mm,            
        bottom=0mm,         
    ]
    \selectfont
    \textbf{\KMAuto} {(specialized models, lack of reasoning)}:
    
    \begin{lstlisting}[language=lean,basicstyle=\ttfamily,frame=none,escapechar=!,breakindent=0pt]
theorem a : IsLeast {x | ∃ n : ℕ, x = n * 7.49 ∧ !\setlength{\fboxsep}{1pt}\fbox{12 $\mid$ n $\wedge$ n $\geq$ 550}!} 344.54 := by sorry
    \end{lstlisting}

    \vspace{-25pt}
    \hspace{238pt}\selectfont\textcolor[RGB]{0, 0, 0}{\fontsize{25}{0}\ding{55}}
    
    \textbf{Mistakes}:  (1) It misunderstands the meaning of dozen (\setlength{\fboxsep}{1pt}\fbox{Box}). (2) The total payment $\textbf{x}$ is not explicitly declared as a \textbf{real number}, causing a type error with the value 7.49.


    \vspace{1pt}
    
    \end{tcolorbox}
\end{minipage}\hfill
\begin{minipage}[t]{0.47\textwidth}
    \vspace{0pt}
    \begin{tcolorbox}[
        colback=white,      
        colframe=black,     
        arc=5pt,            
        boxrule=1pt,        
        width=\linewidth,   
        center,             
        left=0mm,           
        right=0mm,          
        top=0mm,            
        bottom=0mm,         
    ]
    \textbf{\modelname} {\selectfont(ours, formal knowledge + reasoning)}:
    
    \textless think\textgreater

The problem involves math several concepts:

2. **Cost calculation**: Each \fbox{pack of 12} donuts costs \$7.49. The total cost is calculated by multiplying the number of packs by the cost per pack...

Based on the problem, we need to work with the following mathematical objects:

3. **Cost**: \textbf{This is a real number ($\mathbb{R}$)}, calculated as 7.49 × n. We need to find the minimum cost...

Several potential issues might arise when formalizing this problem in Lean:

1. **Type coercion**: For instance, 7.49 is a real number, but we're multiplying it by a natural number n. Lean will handle this through \textbf{automatic coercion}, but we should be aware of it.

...

\textless /think\textgreater
    \begin{lstlisting}[language=lean,basicstyle=\ttfamily,frame=none,escapechar=!,breakindent=0pt]
theorem c : IsLeast {x : ℝ | ∃ n : ℕ, x = n * 7.49 ∧ !\fbox{12 * n $\geq$ 550}!} 344.54 := by sorry
    \end{lstlisting}

    \vspace{-22.5pt}
    \hspace{205pt}\selectfont\textcolor[RGB]{0, 0, 0}{\fontsize{25}{0}\ding{51}}
    
    
    \end{tcolorbox}
\end{minipage}





% ...



% ...



% ...



\end{tcolorbox}
\caption{\textbf{A case study to demonstrate the impact of formal knowledge and informal-to-formal reasoning capability on autoformalization models.} It shows that general-purpose models without formal knowledge make mistakes in code implementation, while specialized ones without reasoning capability struggle with problem understanding and informal-formal alignment. \modelname improves autoformalization performance by combining these two capabilities.} 
\label{fig:bad_case_1}
\end{figure*}

We identify two key abilities for effective autoformalization models:
\textbf{(1) Comprehensive mastery of domain knowledge in formal language.} A model must know how to express every mathematical concept in the target formal language. For instance, without familiarity with the Lean 4 definition of Euler’s totient function, it cannot formalize related problems (see Appendix E.1 in the FormalMATH paper \cite{FormalMath}). Because formal data are scarce in general corpus \cite{Prover-DS-V1} and significant differences exist between versions of formal languages (e.g., Lean 3 and Lean 4 \cite{lean3changes}), general-purpose LLMs often mislearn or overlook crucial language details, limiting their performance (e.g., Claude4-thinking in Figure \ref{fig:bad_case_1}).
\textbf{(2) Reasoning capability of natural language problem understanding and informal-formal alignment.} 
Models tackling real‐world problems must first grasp the intended meaning, which is something specialized models with annotated data in narrow scenarios struggle with (e.g., \KMAuto in Figure \ref{fig:bad_case_1}). Moreover, informal problem statements often omit details (e.g., type hints) that must be made explicit in formal language for rigorous verification. Overcoming these gaps requires strong informal‐to‐formal reasoning abilities, as demonstrated by \modelname (Figure \ref{fig:bad_case_1}) and earlier studies \cite{Beq, mathesis, FMC}. Quantitatively, we leveraged GPT-4o \cite{gpt4o} to classify errors in roughly 10K generated outputs and had human experts annotate a 100-example subset from two autoformalization models (Figure \ref{fig:error_analysis_fewer_infos}, see Appendix \ref{sec:app_analysis} for more details). The results show that \KMAuto suffers a high rate of these two types of errors as an evidence of weak reasoning, whereas our model alleviates these problems.

To equip the model with these two abilities, we introduce \xname, the first data synthesis and training pipeline for the autoformalization model with both domain knowledge and informal-to-formal reasoning capability. We implement this through two key components: (1) large-scale distillation and selection of data with formal knowledge from specialized models, and (2) informal-to-formal reasoning trajectories synthesis guided by an expert-designed autoformalization template.

\begin{figure}
    \centering
    \includegraphics[width=1\linewidth]{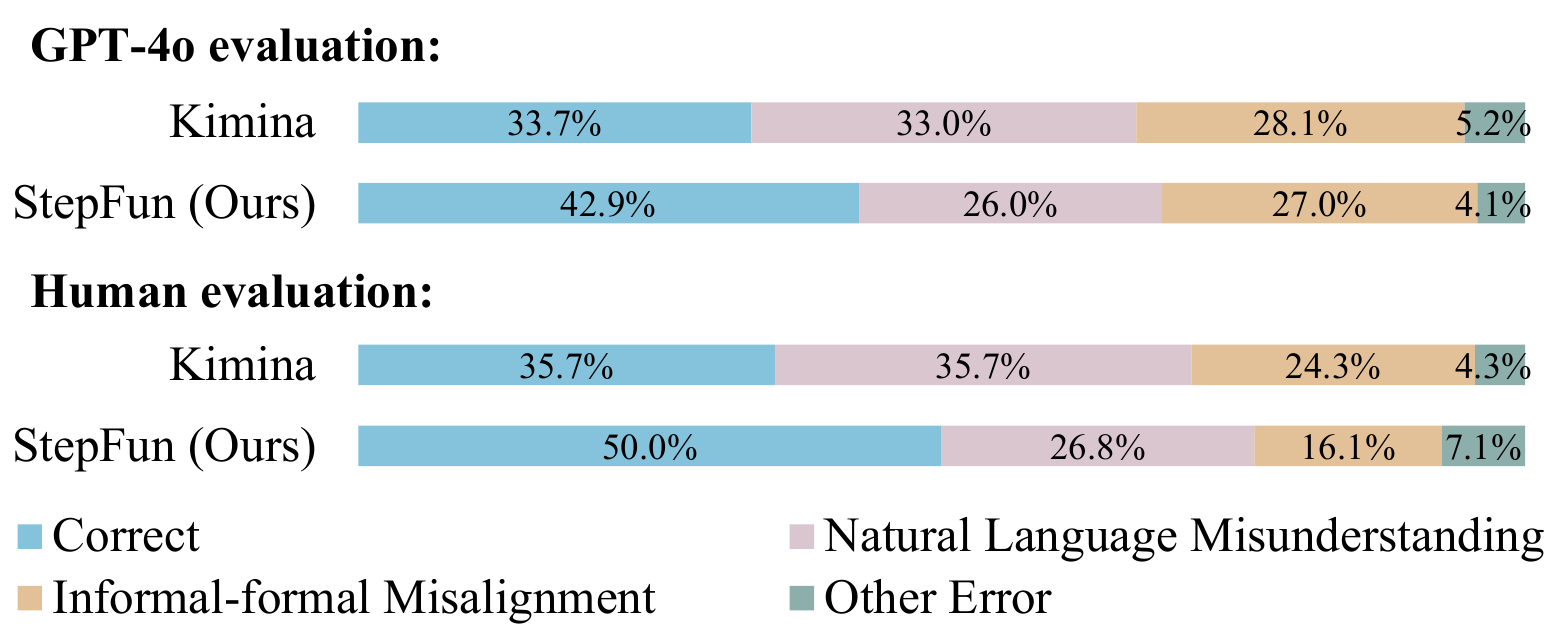}
    \caption{\textbf{Categorical analysis for errors in autoformalization model.} ``Kimina-Auto'' refers to \KMAuto. It illustrates the proportion of two error types in autoformalization, both of which are mitigated in \modelname. }
    \label{fig:error_analysis_fewer_infos}
\end{figure}

Specifically, we first construct two datasets to respectively supplement domain knowledge and enhance the model’s reasoning ability. For domain knowledge, we collect informal-formal data pairs by employing a specialized model (e.g., \KMAuto) to translate a large number of natural language mathematical problems in public datasets (e.g., \NM) into formal statements, followed by majority voting and an LLM-based judge to ensure data quality.
For reasoning ability, we propose a reasoning template that involves problem decomposition and mathematical object mapping. With this template, we leverage a strong instruction-following model (e.g., \Claude) to synthesize the reasoning process for each informal-formal data pair in existing human-annotated datasets.
Next, we use the two datasets to perform supervised fine-tuning on a general-purpose LLM with strong informal mathematical and coding capabilities (e.g. \RRD), thereby integrating domain knowledge and informal-to-formal reasoning ability into a unified model. Finally, we apply reinforcement learning to the fine-tuned model to further promote the fusion of the two capabilities, using the BEq equivalence verification \cite{Beq} as a verifiable reward (RLVR). The overview of \xname is shown in Figure \ref{fig:method_overview}.

Based on this pipeline, we develop two sizes of autoformalization LLMs, \modelname-7B and \modelname-32B. We evaluate them using the BEq verification on mainstream benchmarks, including \FMLite \cite{FormalMath}, \PBench \cite{Prover-DS-V2} and \CBench \cite{CombiBench}. Specifically, \modelname-32B achieves 40.5\% on \FMLite and 26.7\% on \PBench in BEq@1 score \cite{Beq}, setting new SOTA results among both specialized and general-purpose models.




\section{Related Work}
\label{sec:related_work}

\paragraph{Large Language Models for Autoformalization.} Autoformalization is the process of converting mathematical expressions from natural language into their formal language representations \cite{autoformalization_survey}. 
Traditional rule-based autoformalization methods are complex to implement and difficult to generalize \cite{MMA}. Therefore, we mainly focus on LLM-based methods. Earlier works enhance the autoformalization capability of existing LLMs by in-context learning \cite{autoformalizationlargelanguagemodels}, data synthesis with back-translation \cite{MMA, proofnet, inversecoder}, retrieval-augmented generation \cite{Beq}, natural language inference \cite{LeanWorkbook}, and expert iteration with LLM judgers \cite{Prover-Kimina}. Mathesis \cite{mathesis} is the first autoformalization model with reinforcement learning in its training process, but it does not perform informal-to-formal reasoning during translation, resulting in informal-formal misalignment. Moreover, since Mathesis's model and evaluation methods are not publicly available, we are unable to make a performance comparison. For other public works, we compare with them in Section \ref{sec:Experiments}.

The main difference between our method and the aforementioned works is that we equipped the model with both domain knowledge of formal language and informal-to-formal reasoning capabilities, thereby significantly improving its autoformalization capability.

\paragraph{Reasoning-Enhanced Large Language Models.} Inspired by the powerful and effective RL paradigm of general reasoning LLMs like OpenAI-o1~\cite{openaio1}, DeepSeek-R1~\cite{deepseekr1}, and Kimi-K1.5~\cite{kimik1.5}, some previous works attempt to integrate reasoning capabilities from general-purpose LLMs into domain-specific LLMs to enhance their performance in solving complex problems, such as Fin-R1~\cite{Fin-R1}, Table-R1~\cite{tableR1}, CodeV-R1~\cite{zhu2025codev}, and R1-Code-Interpreter~\cite{R1-Code-Interpreter}. 

Since prior works rely on abundant corpora and general LLMs with rich domain knowledge, they typically distil these models to obtain domain-specific reasoning data, but distillation often limits performance below teacher models. In contrast, we first enhance the general model with formal knowledge, then train for reasoning, achieving performance that matches or exceeds both specialized and general models.




\section{Problem Statement}
\label{sec:background}

The autoformalization task discussed in this paper can be formulated as follows:

\begin{definition}[LLM-based Autoformalization]
\label{def:autoformalization}
Given an informal mathematical problem $x$ as the input to an autoformalization model $\mathcal{M}$, its output $\mathcal{M}(x)$ is the corresponding formal statement $y$ with an optional reasoning process. We say that $y$ is a correct formalization of $x$ if and only if:

(1) $y$ passes the syntax check of the formal language.

(2) $y$ is semantically equivalent to $x$.

\end{definition}

Due to the difficulty of automating the strict judgment of semantic equivalence between a model-generated formal statement and an informal problem, we use human-annotated formal statements as the ground truth. We then assess correctness by checking if the model's formalization is equivalent to the human-annotated ground truth using the proof assistant, as in previous work \cite{Beq}:

\begin{definition}[Bidirectional Extended Definitional Equivalence, BEq]
\label{def:beq}
Two formal statements $y_1$ and $y_2$ are bidirectional extended definitional equivalence (denoted as $y_1 \sim y_2$) if and only if there exists a formal proof that derives $y_2$ from $y_1$ using semantics-preserving tactics, and vice versa.
\end{definition}

Weighing both accuracy and computational cost, we use the strictest BEq check, i.e., only calling the symbolic heuristic tactic \texttt{exact?} to automatically search for an equivalence proof between two statements. In our implementation, we use \texttt{sorry} as the proof of $y_1$ to assume its correctness, and use \texttt{exact?} for $y_2$ to search for a proof from $y_1$. If the concatenation of the two proofs passes verification in the Lean 4 REPL and indeed uses $y_1$ in the proof, then $y_2$ is provable from $y_1$. If $y_1$ is also provable from $y_2$, then $y_1 \sim y_2$.
\section{Method}
\label{sec:methods}

\begin{figure*}[h]
    \centering
    \includegraphics[width=1\linewidth]{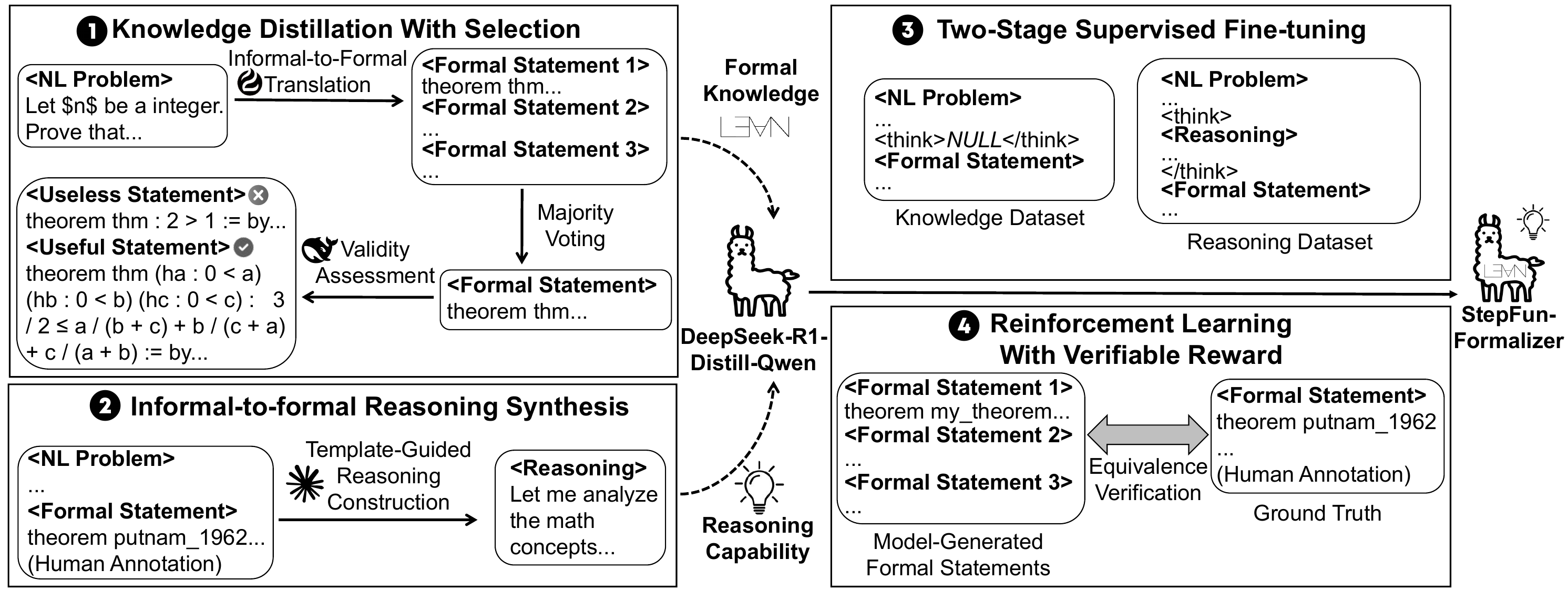}
    \caption{\textbf{The illustration of \xname method.} Our method mainly consists of the construction process for knowledge and reasoning dataset (Section \ref{sec:data_synthesis_1} and \ref{sec:data_synthesis_2}), and the model training process (Section \ref{sec:sft} and \ref{sec:rl}).}
    \label{fig:method_overview}
\end{figure*}

The data synthesis and training pipeline of \xname consists of 4 stages (Figure \ref{fig:method_overview}). 
Stages \circled{1} $\sim$ \circled{2} are used to construct two datasets that enhance the model’s formal knowledge and reasoning capabilities:
\circled{1} \textbf{Knowledge Distillation With Selection.} To obtain a formal knowledge dataset, we need a large number of high-quality informal-formal pairs. Therefore, we employ a specialized LLM to translate informal problems into formal ones, followed by selection to improve the data quality.
\circled{2} \textbf{Informal-to-formal Reasoning Data Synthesis.} Since there is no reasoning process in previous specialized models and distilling reasoning trajectories from general models yields poor results (see Section \ref{sec:ablation_study}), we design a reasoning template for the autoformalization task. Using this template, we synthesize an autoformalization dataset containing informal-to-formal reasoning processes.
Stages \circled{3} $\sim$ \circled{4} involve the training process of our model, which is used to endow the model with the two capabilities and further promote their fusion:
\circled{3} \textbf{Two-Stage Supervised Fine-tuning.} A general-purpose LLM undergoes supervised fine-tuning (SFT) with the two datasets to obtain a model with both capabilities.
\circled{4} \textbf{Reinforcement Learning With Verifiable Reward.} To further facilitate the fusion of knowledge and reasoning, we apply reinforcement learning (RL) to train the fine-tuned model, with BEq checking as a verifiable reward.

\subsection{Knowledge Distillation With Selection}
\label{sec:data_synthesis_1}


\subsubsection{Informal Problem Preparation} Our distillation pipeline begins with the informal mathematical problems $\{x_i\}$ from an open-source dataset, \NM. Following \KMAuto \cite{Prover-Kimina}, we first filter the dataset by manual rules (see Appendix~\ref{sec:app_method_details} for details), resulting in approximately 256K informal problems.
\subsubsection{Formal Statement Generation and Selection}
We prompt \KMAuto to generate 16 candidate formal statements $\{y_{ij}\}_{j=1}^{16}$ for each informal problem $x_i$. Next, we perform a three-tier quality selection on the generated formal statements: 

\textbf{(1) Syntax Check.} We use the Lean4 REPL to perform syntax checking on the formal statements $\{y_{ij}\}_{j=1}^{16}$, and retain the syntactic correct statements, denoted as $\{y^*_{ij}\}_{j=1}^{m_i}$, where $m_i$ is the number of syntactically correct statements.. 

\textbf{(2) Majority Voting.} It is observed that majority voting can significantly improve the performance of autoformalization models (Table \ref{tab:major_voting}), just as it does in coding \cite{alphacode} and informal math \cite{self-consistency} tasks. 

\begin{table}[h]
\fontsize{8}{11}\selectfont
\centering

\begin{tabular}{lccc}
\toprule
Metrics & FormalMATH-Lite & ProverBench & CombiBench \\
\midrule
BEq@1    & 35.1       & 13.3                & 2.6 \\
Maj@16   & \textbf{45.9}       & \textbf{17.2}                & \textbf{3.0}\\
\bottomrule
\end{tabular}%
\caption{
\textbf{Accuracy gains from majority voting for \KMAuto-7B.}
``Maj@16'': generating 16 outputs per problem and selecting one via majority voting for evaluation. See Section \ref{sec:Experiments} for benchmarks and metrics details.}
\label{tab:major_voting}
\end{table}

Specifically, for the syntactically correct formal statements $\{y^*_{ij}\}_{j=1}^{m_i}$, we use BEq verification to partition the formal statements into multiple equivalence classes. Then, one formal statement is randomly selected from the largest group as the optimal formalization $y_i^{**}$ corresponding to the informal problem $x_i$. Formally:
\begin{align*}
    &y_i^{**} = \arg\max_{y^*_{ij}} \sum_{k=1}^{m_i} \indicator (y^*_{ik} \sim y^*_{ij})
\end{align*}

where $\indicator(\cdot)$ is the indicator function, and $\sim$ denotes semantic equivalence. Informal problems and their optimal formalization are collected as a dataset $\{(x_i, y_i^{**})\}$.

\textbf{(3) Problem Validity Assessment.} The selected dataset $\{(x_i, y_i^{**})\}$ is finally evaluated using DeepSeek-V3 (since it is efficient and affordable), removing oversimplified formal statements and those containing inherent contradictions in conditions. We keep approximately 183K informal-formal pairs as the final training data. The ablation study without LLM selection (in Appendix~\ref{sec:app_analysis}) shows that this step can reduce the amount of training data required while maintaining comparable model performance.

\subsection{Informal-to-Formal Reasoning Data Synthesis}
\label{sec:data_synthesis_2}

\begin{figure}[h]
    \centering
    \includegraphics[width=1\linewidth]{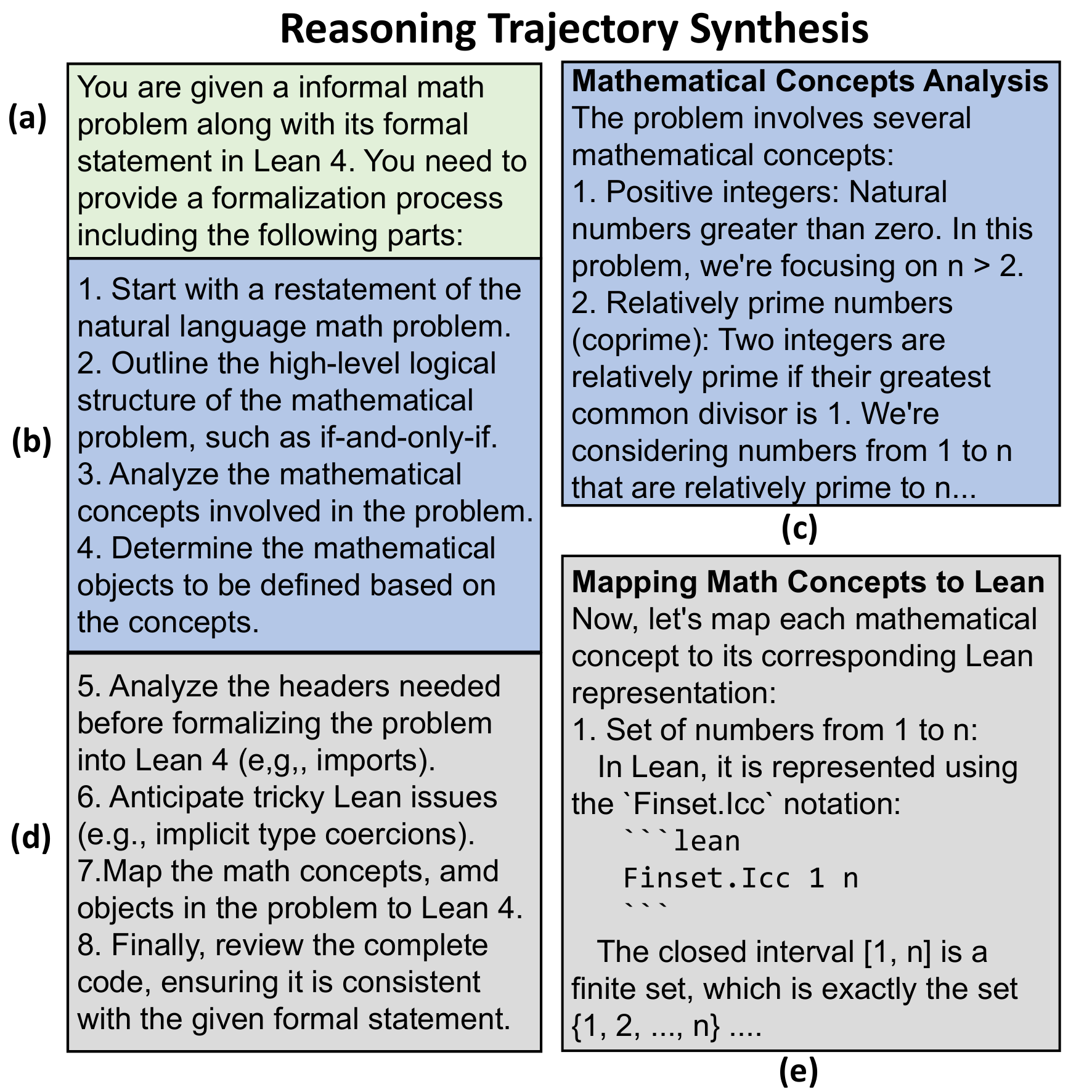}
    \caption{\textbf{The prompts and examples in the template-guided reasoning construction framework.} (a) The task description for autoformalization. (b) Understanding of natural language problems. (c) An example of concept analysis in problem understanding. (d) Analysis of converting informal math objects into formal language. (e) An example of mapping concepts to Lean in informal-to-formal analysis.}
    \label{fig:reasoning_template}
\end{figure}

\subsubsection{Template Design for Reasoning} 

According to the previous error analysis of LLM-based autoformalization, we propose a template-guided reasoning construction framework (Figure~\ref{fig:reasoning_template}), which incorporates the human understanding of the autoformalization process to assist the model in generating reasoning trajectories. The framework consists of two parts:

\textbf{(1) Informal Problem Understanding.} Before delving into the details of formal languages, the model needs to deeply understand the natural language problem, including rephrasing the original question, analyzing its high-level logical structure with the decomposition of the mathematical concepts and corresponding objects involved.

\textbf{(2) Informal-to-Formal Analysis.} To bridge the misalignment between natural and formal language, the model should first consider the tricky issues that may arise during formalization. Then, following a divide-and-conquer paradigm \cite{funcoder}, the model maps the natural language mathematical objects to formal language.

\subsubsection{Synthesizing reasoning trajectories upon existing human-annotated data.} To maximize the correctness of the model's reasoning, we use informal problems with a human-annotated ground-truth formal statement (denoted as $\{(\hat{x}_i, \hat{y}_i)\}$) as seed data. Following previous work \cite{Prover-Kimina, Prover-GD}, the human-annotated data mainly comes from the matched informal-formal statements in automated theorem-proving datasets (see Section \ref{sec:implementation_details} for details). With the annotated data pairs $\{(\hat{x}_i, \hat{y}_i)\}$, we prompt \Claude (since it has strong instruction-following capabilities) to generate a reasoning trajectory $\hat{c}_i$ from $\hat{x}_i$ to $\hat{y}_i$ following our reasoning template. In total, we synthesize 5.8K instances of informal-to-formal reasoning data.

\subsection{Two-Stage Supervised Fine-tuning}
\label{sec:sft}
With the knowledge-distilled dataset $\{(x_i, y_i)\}$ and the reasoning dataset $\{(\hat{x}_i, \hat{c}_i, \hat{y}_i)\}$, we conduct two-stage supervised fine-tuning (SFT) on \RRD, known for its strong reasoning performance in informal mathematics and coding.
Specifically, in the first stage of supervised fine-tuning, $x_i$ is used as the input, and special tokens \texttt{<think></think>} are inserted before the corresponding output $y_i$ to ensure internal format consistency within the model \cite{Qwen3} and maintain its reasoning capability.
During the second-stage supervised fine-tuning, we follow the standard format for reasoning models \cite{deepseekr1} by enclosing the reasoning trajectory $\hat{c}_i$ within \texttt{<think>} and \texttt{</think>} in the output, i.e., \texttt{<think>}$\hat{c}_i$\texttt{</think>}$\hat{y}_i$. After the two-stage supervised fine-tuning, we obtain a preliminary model, \modelname-SFT, equipped with both formal domain knowledge and informal-to-formal reasoning capability.

\subsection{Reinforcement Learning With Verifiable Reward}
\label{sec:rl}

To further enhance the model’s reasoning capability, we perform RL on \modelname-SFT. Due to the lack of high-quality human-annotated data, we use the same set of 5.8K problems $\{(\hat{x}_i, \hat{y}_i)\}$ from the second-stage SFT training for RL training. Despite being used in SFT, continuing RL training on these data still leads to performance improvements (see Section \ref{sec:further_analysis}). The rewards are calculated by performing BEq equivalence verification ($\sim$) between model-generated statements and ground-truth. More formally, the accuracy reward function is defined as follows:
\[
R(y_i, \hat{y}_i) = 
\begin{cases}
1, & \text{if } y_i \sim \hat{y_i} \\
0, & \text{otherwise}
\end{cases}
\]

Taking both training speed and performance into account, we choose Group Relative Policy Optimization (GRPO) algorithm \cite{GRPO} in reinforcement learning training, which eliminates the value function and estimates the advantage in a group-relative manner. 
We also incorporate several improvements from Dynamic Sampling Policy Optimization (DAPO) \cite{DAPO}, including dynamic sampling and token-level loss.

\section{Experiments}
\label{sec:Experiments}

In this section, we present the implementation details of \xname (Sec. \ref{sec:implementation_details}). We conduct a series of experiments (settings in Sec. \ref{sec:exp_settings}) to show the performance comparison between \modelname and previous SOTA models (Sec. \ref{sec:main_results}), alternative designs in our method (Sec. \ref{sec:ablation_study}), the further analysis of RL training and real-world applications (Sec. \ref{sec:further_analysis}).

\subsection{Implementation Details}
\label{sec:implementation_details}

\begin{table}[h]
    \centering
    \fontsize{8}{11}\selectfont
    \begin{tabular}{c|l|c}
    \toprule
                                        & Dataset          & Size \\
    \midrule
    \multirow{5}{*}{Training}   
                                        & MiniF2F \cite{minif2f}         & 488  \\
                                        & ProofNet \cite{proofnet}         & 357 \\
                                        & PutnamBench \cite{PutnumBench}     & 659  \\
                                        & Compfiles \cite{compfiles}       & 115  \\
                                        & FormalMATH-Train \cite{FormalMath} & 5135 \\
                                        \midrule
    \multirow{3}{*}{Evaluation} & FormalMATH-Lite \cite{FormalMath} & 425  \\
                                        & ProverBench \cite{Prover-DS-V2}     & 174  \\
                                        & CombiBench \cite{CombiBench}      & 100  \\
    \bottomrule
    \end{tabular}
    \caption{\textbf{Data partitions and sizes.} We select a subset of the three most recent datasets for evaluation to ensure fairness, with the remaining datasets for training. Specifically, we use a subset of FormalMATH (FormalMATH-Lite) for evaluation, and the remaining (FormalMATH-Train) for training.}
    \label{tab:datasets}
\end{table}

\begin{table*}[h]
\centering

\begin{tabular}{lccccccccc}
\toprule
\multirow{2}{*}{Model} & \multicolumn{2}{c}{FormalMATH-Lite} & \multicolumn{2}{c}{ProverBench} & \multicolumn{2}{c}{CombiBench} \\
&  BEq@1 & BEq@16 & BEq@1 & BEq@16 & BEq@1 & BEq@16 \\
\midrule
\multicolumn{7}{l}{(\textit{General-purpose Models})} \\
OpenAI o3-pro        & 22.6 & 35.5 & 24.7 & 36.2 & \underline{9.0} & 16.0 \\
Claude4-thinking     & 20.8 & 32.2 & 24.4 & 35.6 & \textbf{9.7} & \underline{18.0} \\
Gemini-2.5-thinking  & 17.8 & 31.3 & 20.1 & 36.8 & 8.9 & \underline{18.0} \\
\RR-671B             & 18.4 & 31.3 & 23.5 & 34.5 & 8.1 & \textbf{20.0} \\
DeepSeek-R1-Distill-7B              & 5.2  & 14.6 & 5.4  & 18.4 & 0.4 & 2.0  \\
\midrule
\multicolumn{7}{l}{(\textit{Specialized Models})} \\
LeanFormalizer-PPO-7B           & 18.7 & 24.0 & 12.4 & 18.4 & 0.1 & 1.0 \\
LeanFormalizer-SFT-7B           & 18.9 & 23.3 & 18.4 & 26.4 & 4.8 & 8.0 \\
LeanFormalizer-CoT-7B           & 13.5 & 29.9 & 8.5  & 28.2 & 2.1 & 9.0 \\
Herald-Translator-7B            & 13.6 & 24.7 & 8.2  & 27.0 & 1.3 & 5.0 \\
Goedel-Formalizer-SonnetAnnotated-32B               & 18.7 & 29.2 & 13.6 & 27.6 & 3.4 & 10.0 \\
Goedel-Formalizer-LeanWorkbookAnnotated-32B                   & 15.1 & 26.4 & 5.0  & 12.1 & 0.3 & 2.0 \\
\KMAuto-7B                  & 35.1 & \underline{60.2} & 13.3 & 25.3 & 2.6 & 6.0 \\%
\midrule
\rowcolor[rgb]{0.925,0.925,0.925} \multicolumn{7}{l}{(\textit{Ours})} \\
\rowcolor[rgb]{0.925,0.925,0.925} \modelname-7B        & \underline{38.3}      & \textbf{61.2}  & \underline{25.1}                & \underline{37.9}   & 5.2                       & 11.0 \\
\rowcolor[rgb]{0.925,0.925,0.925} \modelname-32B       & \textbf{40.5}      & 59.3  & \textbf{26.7}             & \textbf{38.5}                  & 6.9              & 14.0 \\
\bottomrule
\end{tabular}%
\caption{BEq@1 and BEq@16 (\%) results of \modelname and baselines on three benchmarks.}
\label{tab:main_results}
\end{table*}

\paragraph{Datasets.} 

The datasets we use for reasoning synthesis, RL training, and evaluation are all collected from automated theorem-proving problem sets (Table \ref{tab:datasets}), which contain informal math problems paired with human-annotated (or model-generated with manually reviewed) formal statements. For problems containing multiple subproblems or lemmas, we retain only the last one. To prevent data contamination, we perform 13-gram decontamination \cite{NGram} and remove training-evaluation overlaps based on problem names.

\paragraph{Training.} We start our training process with \RRD-7B / 32B. In SFT, we train the models for 2 epochs with a learning rate of $2.0 \times 10^{-5}$ and a batch size of 128 in the first stage, and 8 in the second stage. In RL, we use a batch size of 128, a learning rate of $1.0 \times 10^{-6}$, and train 450 steps for the 7B model and 350 steps for the 32B model. The rollout temperature is 1.0. We use the Kimina Lean Server \cite{repl}, equipped with 100 CPU cores and 400 GB of memory, under a 60-second time limit for equivalence verification. The SFT and RL stages are respectively executed on 8 and 32 A800-80G GPUs. The context lengths are 16384. The training of 7B and 32B takes 45.38 hours and 55.85 hours. After training, we obtain \modelname-7B / 32B.  

\subsection{Experimental Settings}
\label{sec:exp_settings}

\paragraph{Benchmarks and Baselines} To show the performance of \modelname, we evaluate it on both in-domain and out-of-distribution (OOD) benchmarks. For in-domain evaluation, we use FormalMATH-Lite \cite{FormalMath}, which shares a similar distribution with our training data (Table \ref{tab:datasets}). For OOD evaluation, we use ProverBench \cite{Prover-DS-V2} and CombiBench \cite{CombiBench}. Some formal statements in these two datasets include additional definitions, functions, and lemmas. We provide them as prompts and ask the model to generate main theorems, to evaluate its generalizability. The benchmarks encompass various areas, including algebra, number theory, and calculus, with difficulty levels ranging from high school to undergraduate.

We compare \modelname with advanced general-purpose LLMs, including o3-pro \cite{o3-pro}, Claude4-thinking \cite{claude4-thinking}, Gemini-2.5-thinking \cite{gemini}, DeepSeek-R1 and \RRD-7B \cite{deepseekr1}. Besides, we also evaluate specialized models, including LeanFormalizer \cite{LeanFormalizer}, Herald Translator \cite{herald}, Goedel-Formalizer \cite{Prover-GD} and \KMAuto \cite{Prover-Kimina}.

\paragraph{Metrics.} We use the BEq@$k$ \cite{Beq} metric to evaluate the models, which is the portion of samples where predicted statements are BEq to ground-truths at least once in $k$ attempts:
\begin{align*}
    &\text{BEq@}k = \frac{1}{N} \sum_{i=1}^{N} \max_{j \in \{1, \cdots, k\}} \indicator (y_{i,j} \sim \hat{y}_{i})
\end{align*}

where $N$ is the sample number; $k$ is the attempt number; $\indicator(\cdot)$ is the indicator function, $\hat{y}_{i}$ is the ground-truth and $y_{i,j}$ is the j-th attempt for the i-th sample. We use BEq@1 to measure the accuracy of the model’s single-shot generation, and BEq@16 to evaluate the upper bound of the autoformalization capability. The temperature is set to 0.6. 

\subsection{Main Results}
\label{sec:main_results}

The evaluation results are shown in Table \ref{tab:main_results}.

\paragraph{Our \modelname model establishes new state-of-the-art results on both FormalMATH-Lite (in-domain) and ProverBench (OOD), demonstrating the effectiveness and generalization of our data synthesis and training pipeline.} Specifically, even \modelname-7B, surpasses every competing model on both benchmarks, offering a computational efficiency advantage: on FormalMATH-Lite it exceeds the previous best, \KMAuto-7B, which serves as its formal knowledge source, and on ProverBench it outperforms large general-purpose models such as \RR-671B. In addition, we observe that the 7B model performs comparably to or slightly better than the 32B model, which may be due to the limited size of the data. The 32B model requires more data for further improvement.

\paragraph{Our model outperforms specialized models with the same parameter scale in CombiBench.} Since modelling combinatorial problems involves complex real-world scenarios and long contexts, it remains very challenging even for advanced general reasoning models to achieve high formalization accuracy on CombiBench, let alone smaller models. After training, our model shows a substantial improvement over specialized models of the same scale, highlighting our model’s capability in real-world scenario understanding.

\subsection{Design Alternatives of Our Method}
We explore design alternatives of our method, including the ablation study of the knowledge and reasoning datasets, another reasoning data collection method, and a different base model (see Appendix \ref{sec:app_analysis}). To highlight how each choice impacts generalizable performance, we evaluate them on OOD benchmarks.
\label{sec:ablation_study}

\paragraph{Both the knowledge and reasoning datasets contribute to the improvement of autoformalization.} 

We conduct ablation studies to investigate the individual contributions of these two parts (knowledge and reasoning) of our data:
\modelname-7B is re-trained with the same approach except removing the first stage (knowledge) and the second stage (reasoning) of SFT (Section \ref{sec:sft}) separately. The results (Table~\ref{tab:ablation_study_1}) show that informal-to-formal reasoning is the main contributor to model performance, while formal knowledge serves a complementary role. Notably, removing reasoning data leads to a sharp drop in BEq@16, highlighting its importance in boosting the performance upper bound.

\begin{table}[h]
\fontsize{8}{11}\selectfont
\setlength{\tabcolsep}{1.0mm}
\centering

\begin{tabular}{lcccc}
\toprule
\multirow{2}{*}{Datasets} & \multicolumn{2}{c}{ProverBench} & \multicolumn{2}{c}{CombiBench} \\
&  BEq@1 & BEq@16 & BEq@1 & BEq@16 \\
\midrule
\rowcolor[rgb]{0.925,0.925,0.925} \xname (Ours)   & \textbf{25.1}       & \textbf{37.9}    & \underline{5.2}  & \textbf{11.0} \\
w/o Knowledge   & \underline{24.5}    & \underline{37.4} & 3.9              & \underline{10.0} \\
w/o Reasoning   & 21.8                & 25.3             & \textbf{5.3}              & 6.0 \\
\bottomrule
\end{tabular}%
\caption{Comparison of the roles of the two training datasets of the SFT stage.}
\label{tab:ablation_study_1}
\end{table}

\paragraph{The designed reasoning template helps the model better perform informal-to-formal translation.} To show the effectiveness of the reasoning template, we replace the reasoning data in SFT with reasoning trajectories directly distilled from a general-purpose LLM. Specifically, we prompt Claude4-thinking \cite{claude4-thinking}, a reasoning model of the same series as Claude 3.7 Sonnet, which we used to synthesize reasoning trajectories, to translate the problems in annotated datasets into formal statements, and use BEq to select the correct translations along with reasoning. The training results are shown in Table \ref{tab:reasoning_template_vs_claude4}. It indicates that the reasoning process produced by Claude4-thinking causes a significant decline in the model’s performance. We observe that, during formalization, the general reasoning model devotes its efforts to \emph{solving} the informal problem instead of \emph{formalizing} it, and thereby constraining its overall learning capability (refer to Appendix~\ref{sec:app_analysis} for examples).

\begin{table}[h]
\centering
\setlength{\tabcolsep}{1.0mm}
\fontsize{8}{11}\selectfont

\begin{tabular}{lcccc}
\toprule
\multirow{2}{*}{Method} & \multicolumn{2}{c}{ProverBench} & \multicolumn{2}{c}{CombiBench} \\
&  BEq@1 & BEq@16 & BEq@1 & BEq@16 \\
\midrule
\rowcolor[rgb]{0.925,0.925,0.925} Template (Ours) & \textbf{25.1}       & \textbf{37.9}    & \textbf{5.2}  & \textbf{11.0} \\
Direct Distillation & 21.8 & 33.3 & 4.8  & 10.0 \\
\bottomrule
\end{tabular}%
\caption{Comparison between template-guided reasoning trajectory synthesis and direct distillation from a general-purpose reasoning model (Claude4-thinking).}
\label{tab:reasoning_template_vs_claude4}
\end{table}


\subsection{Further Analysis}
\label{sec:further_analysis}

\paragraph{Reinforcement learning can consistently improve the model’s autoformalization capability.} 

To show the performance improvement brought by reinforcement learning, we evaluate \modelname-7B every 50 training steps and record the average BEq@1 of all benchmarks. The relationship curve between BEq@1 and the training reward is shown in Appendix \ref{sec:app_analysis}. As training progresses, both the reward and downstream task performance improve in tandem. After 450 training steps, the reward increases from 0.232 to 0.347, and the average BEq@1 improves from 0.258 to 0.303, which underscores the effectiveness of reinforcement learning.

\paragraph{Our model generates more verifiable formal statements in different domains.} 



\begin{figure}[h]
    \centering
    \includegraphics[width=1\linewidth]{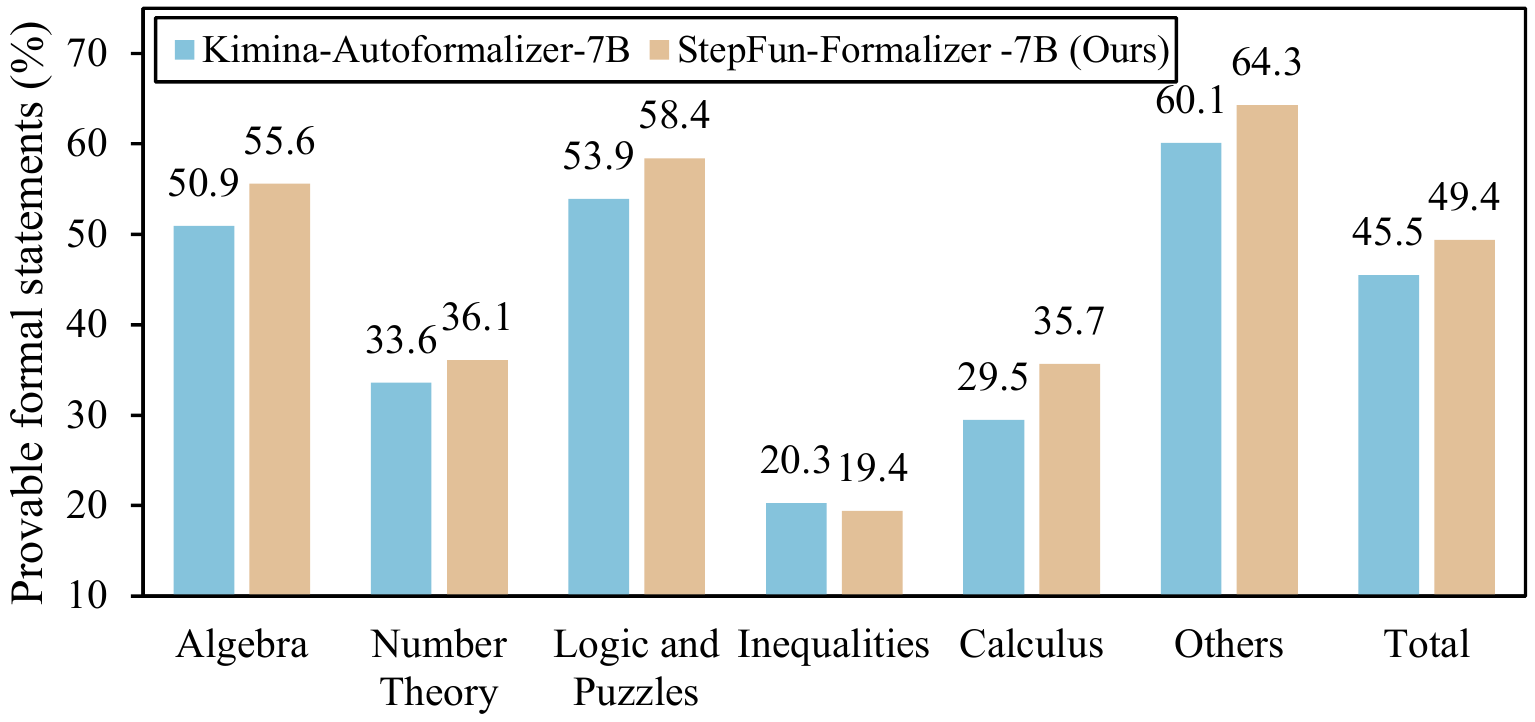}
\caption{The proportion (\%) of provable formal statements among 10K problems from \NM in each domain.}
\label{tab:proved_statements}
\end{figure}

We conduct a simulation experiment to show the performance of \modelname in end-to-end theorem proving from natural language \cite{mathesis}. Specifically, we randomly select 10K problems in \NM, translate them into formal statements using \modelname-7B and \KMAuto-7B, and then use Kimina-Prover-7B \cite{Prover-Kimina} to generate 16 proofs for each statement. The results show that Kimina-Prover proves 4940 formal statements from \modelname-7B and 4549 from \KMAuto-7B. Detailed statistics of provable statements in each domain are shown in Figure \ref{tab:proved_statements}. Our model generates a higher proportion of provable formal statements across all domains except inequalities, highlighting its effectiveness in end-to-end theorem proving.

\section{Conclusion}
\label{sec:Conclusion}
In this paper, we propose \xname, a data synthesis and training pipeline for LLM-based autoformalization. The pipeline mitigates the lack of formal knowledge in general-purpose models via large-scale distillation and selection, and improves the model’s understanding of natural language problems through template-guided reasoning synthesis, which facilitates complex informal-to-formal translation. We train 7B and 32B models with this pipeline and evaluate them on three benchmarks. Experimental results demonstrate that \modelname outperforms both general-purpose and specialized models on FormalMATH-Lite and ProverBench. We also conduct additional analyses to investigate the roles of each component in the pipeline.
\section*{Acknowledgements}
\label{sec:acknowledgements}

This work is partially supported by the Strategic Priority Research Program of the Chinese Academy of Sciences (Grants No.XDB0660300, XDB0660301, XDB0660302), Science and Technology Major Special Program of Jiangsu (Grant No. BG2024028), the NSF of China (Grants No. 62341411, 62222214, 6240073476), CAS Project for Young Scientists in Basic Research (YSBR-029) and Youth Innovation Promotion Association CAS.

\FloatBarrier
\bibliography{aaai2026}

\FloatBarrier
\clearpage
\appendix
\section{Implementation Details}
\label{sec:app_method_details}

\subsection{Informal Problem Preparation}
\label{sec:app_informal_problem_prep}

To obtain high-quality informal mathematical problems for knowledge distillation (Section \ref{sec:data_synthesis_1}), we use manual rules to filter the problems in \NM \cite{NuminaMath}. Specifically, we retain data samples that satisfy the following requirements:

\begin{enumerate}
    \item The problem needs to be complete and should not contain multiple sub-questions, which can be determined by the \texttt{problem\_is\_valid} field.
    \item The \texttt{problem\_type} field should not be \texttt{Geometry} or \texttt{Combinatorics}, since these two types of problems are still challenging to formalize for existing LLMs.
    \item The problem is a proof problem, or is a calculation problem with an answer containing only numbers, commas, and parentheses. For the latter, we prepend ``Show that it is" to the answer and append it to the problem, thereby transforming it into a proof problem.
\end{enumerate}

After filtering, we collect 256K informal problems from the 896K data in \NM that are suitable for formal knowledge distillation. We present the distribution of problem types before and after filtering in Section \ref{sec:distribution_of_problem_types}.

\subsection{Dataset Construction}

\begin{figure}[h]
    \centering
    \includegraphics[width=1\linewidth]{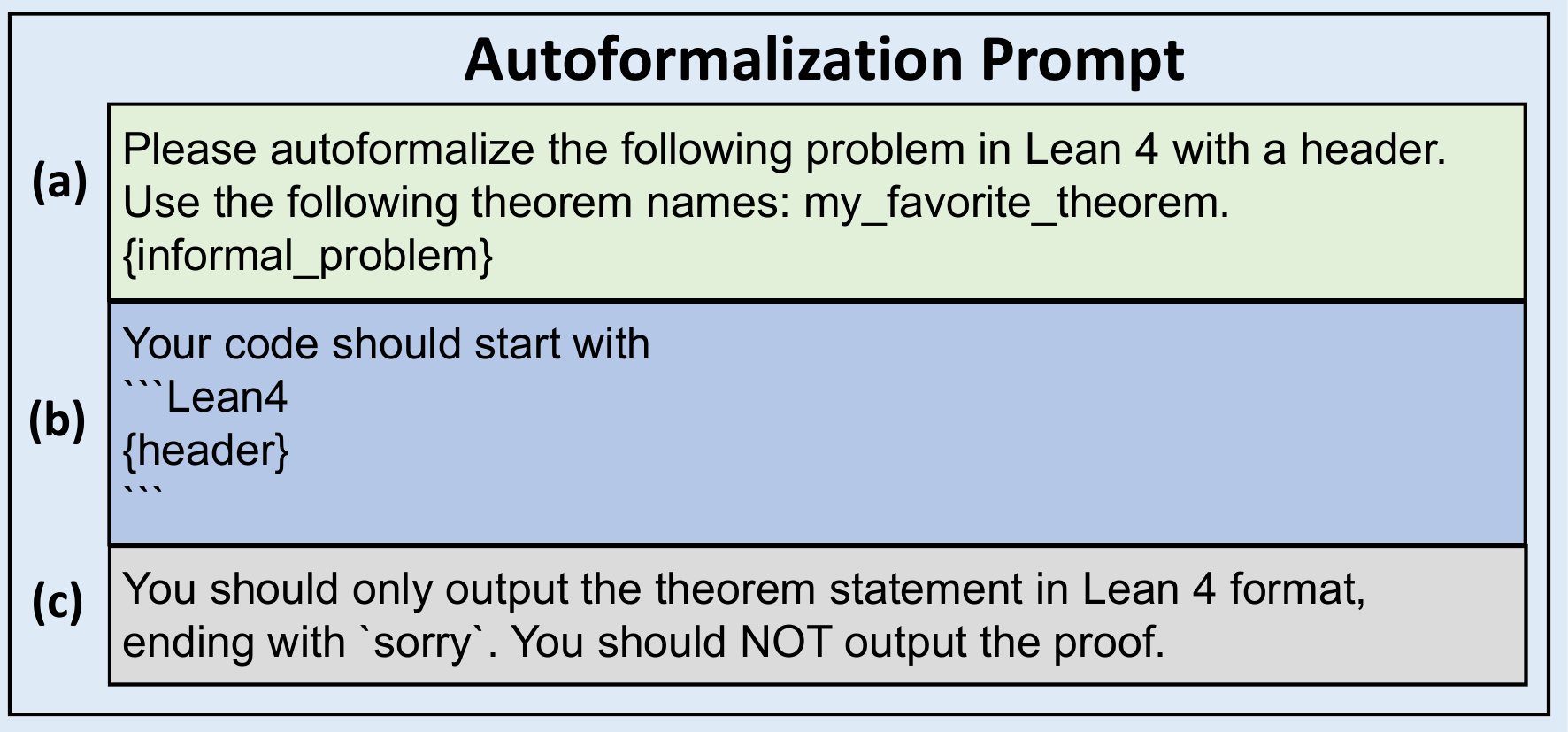}
    \caption{\textbf{The prompt for autoformalization.} (a) The original prompt of \KMAuto. (b) The prompt used to indicate which libraries need to be imported, or which predefined functions or lemmas are available for use. (c) Additional requirements to prevent the model from generating a formal proof.}
    \label{fig:prompt_autoformalization}
\end{figure}

In the knowledge distillation stage (Section \ref{sec:data_synthesis_1}), we follow \KMAuto \cite{Prover-Kimina}, using the same prompts (Figure~\ref{fig:prompt_autoformalization}(a)) and inference hyperparameters. In the reasoning trajectory synthesis stage (Section \ref{sec:data_synthesis_2}), we use the reasoning template (Figure~\ref{fig:prompt_reasoning_synthesis}, Page 15) with few-shot examples (Figure~\ref{fig:prompt_reasoning_synthesis_few_shot_1}$\sim$\ref{fig:prompt_reasoning_synthesis_few_shot_3}, Page 16$\sim$21) to prompt \texttt{claude-3-7-sonnet-20250219} for reasoning synthesis. We set the temperature to 0.6, and the maximum context length is 16384.

\subsection{Training}

\paragraph{Hyperparameters.} For SFT, we perform full-parameter fine-tuning using DeepSpeed ZeRO-3 for parallel, AdamW as the optimizer, and cosine decay as the learning rate scheduling strategy. We do not use warm-up or packing. The numerical precision of our model is BF16. For RL, we sample 32 data points at each step, and we generate 32 rollouts for each sample. The coefficient for the KL divergence is set to $1 \times 10^{-4}$. We use a linear scheduler for the learning rate. Other important hyperparameters can be found in Section~\ref{sec:implementation_details}.

\paragraph{Prompts.} For knowledge datasets, we use the same prompt as \KMAuto (Figure~\ref{fig:prompt_autoformalization}(a)). For reasoning datasets and RL training, we add prompts with headers in training data (Figure~\ref{fig:prompt_autoformalization}(b)) to align with the evaluation of OOD tasks, which may provide additional definitions or lemmas for the model to apply.

\subsection{Evaluation}

\paragraph{Hyperparameters.} We align the hyperparameters of the evaluated specialized models with those they released, if available. For the general-purpose models and \modelname, we set the temperature to 0.6 and the maximum context length to 16384.

\paragraph{Prompts.} For a specialized model, if its training prompt has been released, we use the released prompt for evaluation, such as Goedel-Formalizer \cite{Prover-GD}. Otherwise, we use the prompt in Figure~\ref{fig:prompt_autoformalization}(a). Following EvalPlus \cite{EvalPlus}, we prepend the header in benchmarks as the response prefix for specialized models to maximize their performance. For general-purpose models, since some of them do not understand the autoformalization task well (e.g., \RRD-7B), we add additional instructions (Figure~\ref{fig:prompt_autoformalization}(c)) to prevent them from generating a formal proof. For \modelname, we keep the evaluation prompt consistent with RL training, i.e. Figure~\ref{fig:prompt_autoformalization}(a)(b).

\subsection{BEq Verification}

\paragraph{Using an additional theorem-proving model to check the equivalence of formal statements incurs significant computational overhead, while providing only marginal accuracy gains.} We randomly sample 320 formal statements generated by the model and conduct equivalence checking with the ground truth using both Lean4's built-in search tactic \texttt{exact?} and InternLM-Math-Plus-20B \cite{internlm-math}, as in the BEq paper \cite{Beq}. The results show that the latter identified only 3 more correct formal statements than the former, while taking approximately 4 times longer time cost (3 min vs 12 min) and requiring additional GPU resources (CPU only vs 8 A100-80G GPUs). Considering the computational cost of RL training and large-scale evaluation, we choose to use only \texttt{exact?} for equivalence checking.

\section{Further Analysis}
\label{sec:app_analysis}

\subsection{Categorical Analysis for Errors in Autoformalization Models}

\begin{figure*}
    \centering
    \begin{minipage}[t]{0.49\linewidth}
        \centering
        \includegraphics[width=\linewidth]{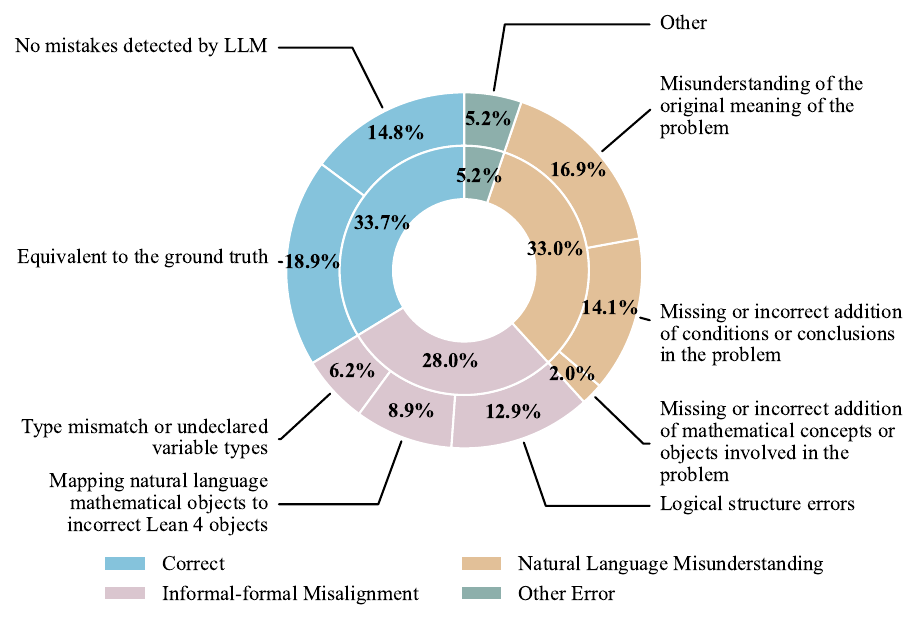}
        \caption*{(a) \KMAuto-7B}
    \end{minipage}
    \hfill
    \begin{minipage}[t]{0.49\linewidth}
        \centering
        \includegraphics[width=\linewidth]{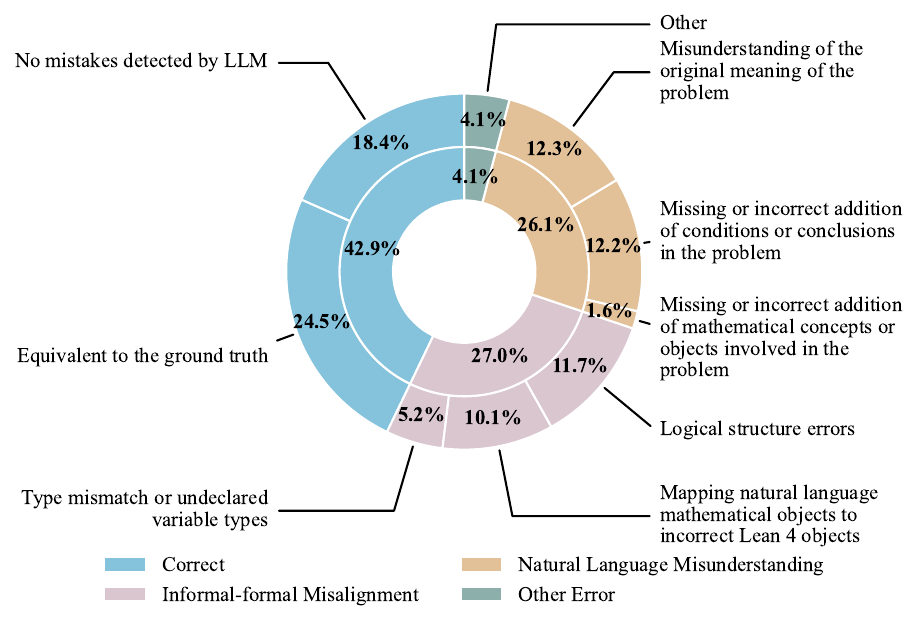}
        \caption*{(b) \modelname-7B}
    \end{minipage}
    \caption{Categorical analysis for errors in two models.}
    \label{fig:error_analysis}
\end{figure*}

To identify the key factors affecting the performance of the autoformalization model (Section \ref{sec:introduction}), we employ two human experts in Lean 4 to conduct error analysis on 100 model-generated formal statements. The most common error types identified by the experts in their analysis are detailed as follows:

\begin{enumerate}
    \item Misunderstanding of the original meaning of the problem.
    \item Missing or incorrect addition of mathematical concepts or objects involved in the problem (e.g., the constant was missing in indefinite integrals).
    \item Missing or incorrect addition of conditions or conclusions in the problem (e.g., the verification of the equality condition was missing in inequality problems).
    \item Mapping natural language mathematical objects to incorrect Lean 4 objects.
    \item Logical structure errors (e.g., the informal problem implies a ``if and only if'' relationship, but the formal statement only includes one direction).
    \item Type mismatch or undeclared variable types (e.g., the problem does not specify the type of a mathematical object, but this type must be explicitly stated in the formal language; otherwise, a syntax error will occur).
\end{enumerate}

We further abstract the first three types of errors as \textbf{Natural Language Misunderstanding}, where the model misinterprets certain natural language concepts \emph{before} converting them into formal language. The last three types of errors are summarized as \textbf{Informal-Formal Misalignment}, where the natural language problem is correctly understood, but mistakes in formal language occur \emph{during} the translation process.

\begin{figure}
    \centering
    \includegraphics[width=1\linewidth]{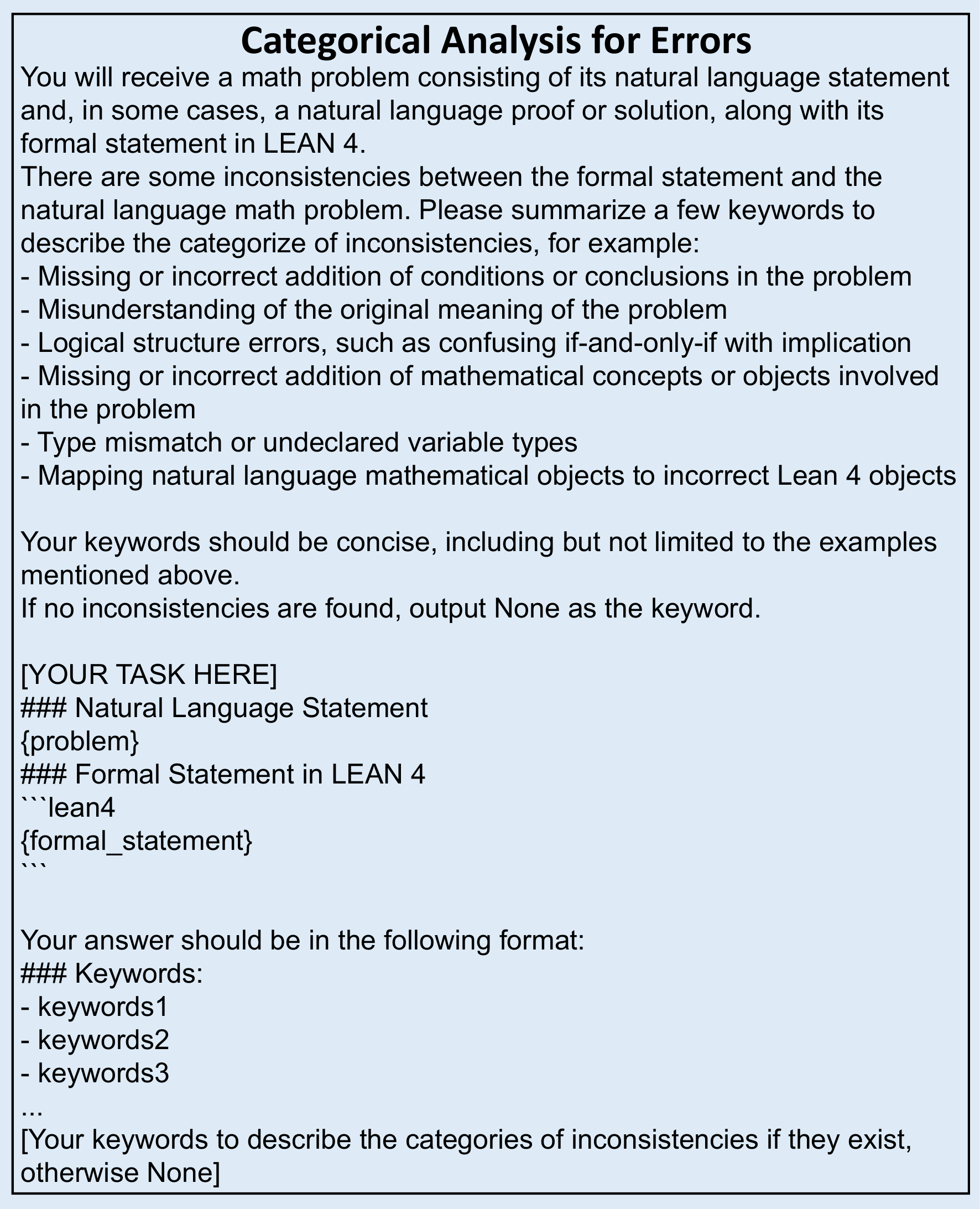}
    \caption{The prompt for LLM-based error type analysis.}
    \label{fig:prompt_error_analysis}
\end{figure}

Then, we used GPT-4o to conduct error analysis on roughly 10K formal statements generated by \KMAuto \cite{Prover-Kimina} and \modelname across three benchmarks. Based on the analysis, we identified the distribution of two main errors. The detailed statistical results are shown in Figure~\ref{fig:error_analysis}. It demonstrates that these two types of errors can be mitigated by \modelname. The evaluation prompts are elaborated in Figure \ref{fig:prompt_error_analysis}.

\subsection{Knowledge Data Selection by LLM}
\label{sec:app_llm_data_selection}
\begin{figure}[h]
    \centering
    \includegraphics[width=1\linewidth]{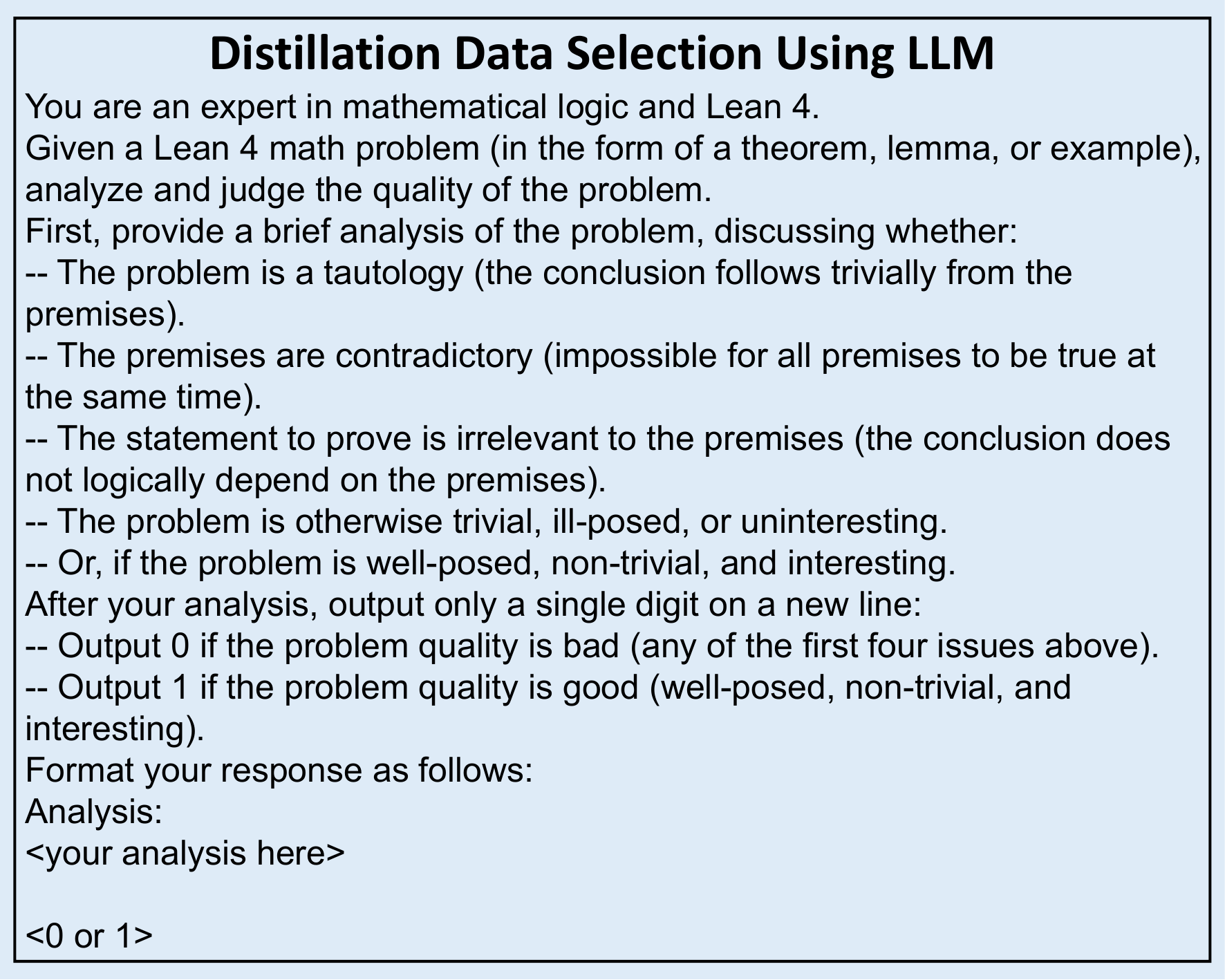}
    \caption{The prompt for LLM-based data selection.}
    \label{fig:prompt_data_selection}
\end{figure}

\paragraph{LLM-based data selection in knowledge distillation can improve training efficiency without compromising performance.} In Section \ref{sec:data_synthesis_1}, we use DeepSeek-V3 to conduct validity filtering on the distilled formal statements. Specifically, we remove invalid problems due to tautology, contradictions, irrelevance, or triviality. The prompt is shown in Figure \ref{fig:prompt_data_selection}. After LLM selection, the total data volume decreased from 253913 to 182548. The distribution of problem types before and after LLM selection is shown in Section \ref{sec:distribution_of_problem_types}.

To verify the effectiveness of LLM selection, we conduct an additional ablation experiment to compare the training performance (Table \ref{tab:ablation_data_selection_perf}), with and without LLM selection. The results show that using LLM selection can reduce the amount of training data with minimal impact on performance.

\begin{table}[h]
\fontsize{8}{11}\selectfont
\setlength{\tabcolsep}{1.0mm}
\centering

\begin{tabular}{lcccc}
\toprule
\multirow{2}{*}{Method} & \multicolumn{2}{c}{ProverBench} & \multicolumn{2}{c}{CombiBench} \\
& BEq@1 & BEq@16 & BEq@1 & BEq@16 \\
\midrule
\rowcolor[rgb]{0.925,0.925,0.925} \xname (Ours)   & 25.1       & \textbf{37.9}    & \textbf{5.2}  & 11.0 \\
w/o LLM selection   & \textbf{25.4}    & 35.6   & 4.9              & \textbf{13.0} \\
\bottomrule
\end{tabular}%
\caption{Performance comparison of with and without LLM selection in knowledge data distillation.}
\label{tab:ablation_data_selection_perf}
\end{table}

\subsection{Distribution of Problem Types}
\label{sec:distribution_of_problem_types}

\begin{figure}[h]
    \centering
    \includegraphics[width=1\linewidth]{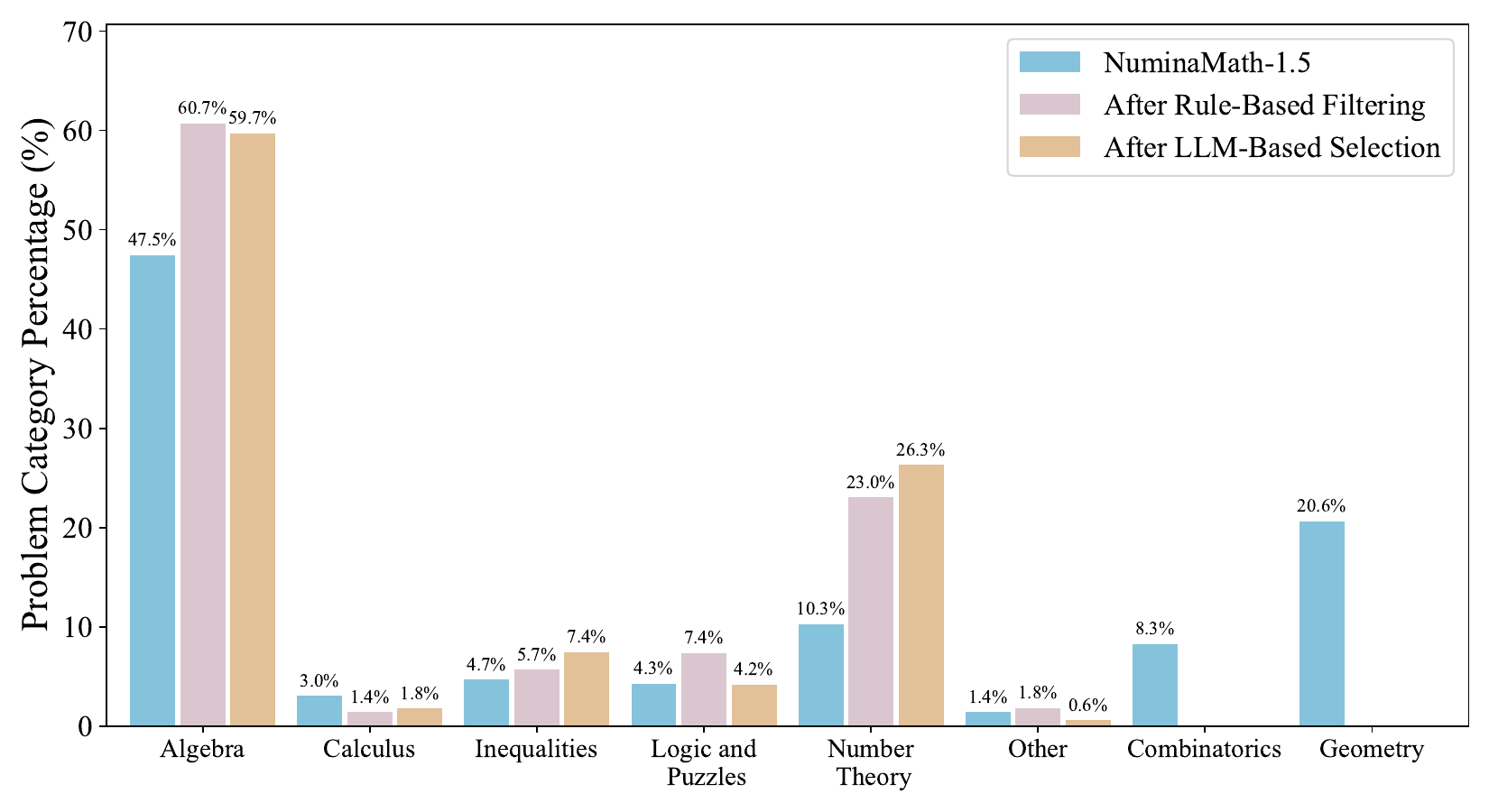}
    \caption{Comparison of problem type distributions.}
    \label{fig:problem_types_after_filter}
\end{figure}

We report the proportions of problem types in datasets for (1) original 896K informal problems in \NM, (2) 256K subset filtered by manual rules, and (3) 183K subset with formal statements further selected via LLM-based methods. The distribution of problem types is shown in Figure \ref{fig:problem_types_after_filter}.
It shows that the rule-based method tends to filter out more algebra and number theory problems, which are easier to formalize. After LLM-based selection, the proportion of logic and puzzle problems decreases, suggesting that some of the formalizations in this category are invalid.

\subsection{Alternative Base Models}
\paragraph{Training from a general-purpose reasoning model helps improve the model’s formalization capability.} We replace the base model from \RRD-7B to \KMAuto-7B in training, which possesses more comprehensive formal knowledge rather than general reasoning capabilities. The results (Table \ref{tab:base_models}) show that a base model with strong math and coding reasoning capabilities also contributes to the autoformalization process.

\begin{table}[h]
\centering
\setlength{\tabcolsep}{0.8mm}
\fontsize{8}{11}\selectfont

\begin{tabular}{lcccc}
\toprule
\multirow{2}{*}{Models} & \multicolumn{2}{c}{ProverBench} & \multicolumn{2}{c}{CombiBench} \\
&  BEq@1 & BEq@16 & BEq@1 & BEq@16 \\
\midrule
\rowcolor[rgb]{0.925,0.925,0.925} \modelname-R1D (Ours) & \textbf{25.1} & \textbf{37.9}  & \textbf{5.2} & \textbf{11.0} \\
\modelname-KM  & 21.4 & 30.5 & 4.3  & 8.0 \\
\bottomrule
\end{tabular}%
\caption{The impact of different base models on the training performance of \modelname. ``R1D'': \RRD-7B; ``KM'': \KMAuto-7B.}
\label{tab:base_models}
\end{table}

\subsection{Training Curve in Reinforcement Learning}

\begin{figure}[h]
    \centering
    \includegraphics[width=\linewidth]{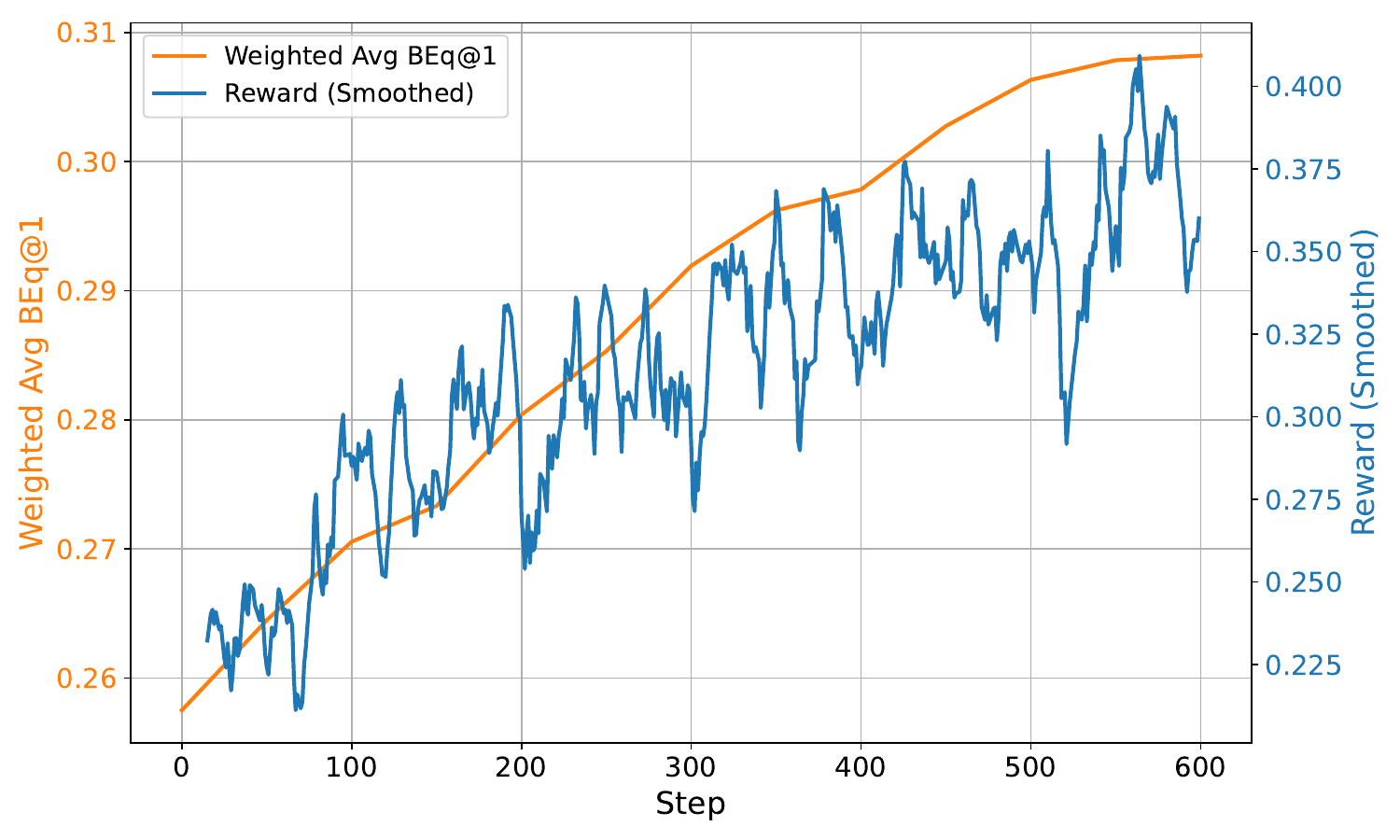}
    \caption{The performance curves of reward and weighted average BEq@1 across all benchmarks.}
    \label{fig:reward_and_beq_plot}
\end{figure}

RL training curves of the reward and average BEq@1 on three benchmarks are shown in Figure~\ref{fig:reward_and_beq_plot}. We observe that both the reward and the performance of downstream tasks increase as RL progresses, which demonstrates the effectiveness of RL training.

\subsection{Off-Task Behaviours in General Reasoning LLMs During Formalization}

We observe that general-purpose reasoning models like Claude4-thinking tend to exhibit off-task behaviour during the formalization reasoning process. Specifically, they devote much of their reasoning to \emph{solving} the informal problem rather than \emph{formalizing} it, yet still manage to produce the correct formal statement. Figure \ref{fig:claude4_bad_example} (Page 22) shows an example.
This phenomenon suggests that directly distilling general reasoning models to obtain autoformalization reasoning trajectories may harm the model's performance (Section \ref{sec:ablation_study}). That is why we use a manually designed reasoning template for reasoning data synthesis (Section \ref{sec:data_synthesis_2}).

\subsection{Applications: Data synthesis for Training Theorem-proving Models}
\label{sec:application_in_prover}

\paragraph{Our model can be used to synthesize more diverse verifiable formal statements, thereby facilitating the training of theorem-proving models.} Inspired by Goedel-Formalizer \cite{Prover-GD}, we attempt to combine the formalization capabilities of \modelname and \KMAuto to provide more diverse training data for theorem-proving models.

We conduct an experiment to simulate the process of distilling the theorem-proving ability from a larger model into a smaller one. Specifically, we employ \modelname and \KMAuto to formalize the 256K problems from NuminaMath-1.5, and then employ Kimina-Prover-7B to generate the corresponding formal proofs. We merge and \emph{deduplicate} the data from two models and verify the proofs by Lean 4 REPL. The verified proofs are used to fine-tune Qwen2.5-Math-1.5B-Instruct, which is the base model of Kimina-Prover-1.5B. We use FormalMATH-Lite \cite{FormalMath} and CombiBench \cite{CombiBench} to evaluate the theorem-proving performance of the models fine-tuned with 4 different kinds of synthetic datasets (Table \ref{tab:training_for_prover}):

\begin{table}[h]
\fontsize{8}{11}\selectfont
\setlength{\tabcolsep}{0.9mm}
\centering

\begin{tabular}{l|c|cccc}
\toprule
\multirow{2}{*}{Data Source} & \multirow{2}{*}{Data Size} & \multicolumn{2}{c}{FormalMATH-Lite} & \multicolumn{2}{c}{CombiBench} \\
& & Pass@1 & Pass@32 & Pass@1 & Pass@32 \\
\midrule
Kimina-Prover-1.5B & Unknown & 30.2    & 43.5   & 1.0    & \textbf{4.0} \\
Kimina & 89473 & 31.4    & 44.9   & 1.0    & 2.0 \\
Kimina $\times$ 2 & 112121 & \textbf{32.5}    & 45.4   & 1.0    & 3.0 \\
\rowcolor[rgb]{0.925,0.925,0.925} Kimina + Ours & \textbf{120602} & \textbf{32.5}    & \textbf{46.4}   & \textbf{1.8}    & \textbf{4.0} \\
\bottomrule
\end{tabular}%
\caption{Performance (\%) comparison between theorem-proving models trained on different distilled datasets.}
\label{tab:training_for_prover}
\end{table}

\begin{enumerate}
    \item \textbf{Kimina-Prover-1.5B}: The official model, distilled from Kimina-Prover-72B.
    \item \textbf{Kimina}: Generating one formal statement using \KMAuto.
    \item \textbf{Kimina $\times$ 2}: Generating two formal statement using \KMAuto with deduplication.
    \item \textbf{Kimina + Ours}: One formal statement from \KMAuto, and another formal statement from \modelname, followed by deduplication.
\end{enumerate}

The results demonstrate that the combined dataset generated from \modelname and \KMAuto is the largest under the same computation cost, and the model trained on it achieves the best performance, even surpassing the official model, which implies that our model can generate more diverse verifiable formal statements, thereby promoting the development of theorem-proving models.

\subsection{Distribution Differences Across Benchmarks}

Since we use a subset of FormalMATH~\cite{FormalMath} (FormalMATH-Lite) for evaluation and the remaining portion for training, we consider FormalMATH-Lite as an in-domain benchmark. To highlight the differences between FormalMATH-Lite and ProverBench~\cite{Prover-DS-V2}, we compute the n-gram perplexities within each dataset and between the two datasets:

\begin{table}[h]
\centering

\begin{tabular}{llc}
\toprule
Training Dataset & Test Dataset   & Perplexity \\
\midrule
FormalMATH-Lite & FormalMATH-Lite & 12.31    \\
FormalMATH-Lite & ProverBench     & 436.20   \\
ProverBench     & ProverBench     & 12.11    \\
ProverBench     & FormalMATH-Lite & 468.00   \\
\bottomrule
\end{tabular}%
\caption{N-gram perplexities within and between datasets.}
\label{tab:n_gram_ppl}
\end{table}

Specifically, we train an n-gram model on the dataset in the first column (Training Dataset) in Table~\ref{tab:n_gram_ppl} and compute the average perplexity on the dataset in the second column (Test Dataset). The results show that the average OOD ratio between FormalMATH-Lite and ProverBench is 37.04, indicating a large distributional difference.

\subsection{Comparison with Goedel-Formalizer-V2}

We compare \modelname with the concurrent work Goedel-Formalizer-V2 (2025-08-05)~\cite{Prover-GD-V2}, which distils 50K reasoning data from Claude Sonnet 4, at a cost substantially higher than that of our data synthesis method. The evaluation results are shown in Table~\ref{tab:compare_with_goedel}:

\begin{table}[h]
\centering
\setlength{\tabcolsep}{0.8mm}
\fontsize{8}{11}\selectfont

\begin{tabular}{lcccc}
\toprule
\multirow{2}{*}{Models} & \multicolumn{2}{c}{ProverBench} & \multicolumn{2}{c}{CombiBench} \\
&  BEq@1 & BEq@16 & BEq@1 & BEq@16 \\
\midrule
\rowcolor[rgb]{0.925,0.925,0.925} \modelname-7B (Ours) & \textbf{25.1} & \textbf{37.9}  & \textbf{5.2} & \textbf{11.0} \\
Goedel-Formalizer-V2-8B  & 16.3 & 28.7 & 3.3  & 10.0 \\
\rowcolor[rgb]{0.925,0.925,0.925} \modelname-32B (Ours)  & \textbf{26.7}             & \textbf{38.5}                  & \textbf{6.9}              & \textbf{14.0} \\
Goedel-Formalizer-V2-32B  & 15.6 & 28.2 & 4.1  & 13.0 \\
\bottomrule
\end{tabular}%
\caption{Performance (\%) comparison between our model and Goedel-Formalizer-V2.}
\label{tab:compare_with_goedel}
\end{table}

We note that Goedel-Formalizer-V2's training prompt does not include headers required in BEq, which causes difficulties for our evaluation. Since it is a reasoning model, we cannot simply take headers as prefixes in its output, as is done for Kimina-Autoformalizer. Although we attempt to add headers to its prompts, the model's limited instruction-following ability prevents it from generating specified headers. As a result, its BEq scores may be underestimated. Some works note this issue and replace BEq with LLM-as-a-judge, but it may introduce additional bias due to the randomness of LLM's outputs. We will refine our evaluation methodology in future work.

\section{Case Study}

A detailed case study of \modelname’s comprehensive reasoning trajectories is presented in Figure \ref{fig:case_study_1} (Page 23$\sim$24).

\begin{figure*}[t]
    \centering    
    \begin{tcolorbox}[
        colback=white,      
        colframe=white,     
        arc=0pt,            
        boxrule=1pt,        
        width=\linewidth,
        center,             
        left=0mm,           
        right=0mm,          
        top=0mm,            
        bottom=0mm,         
    ]
    \begin{tcolorbox}[
        colback=white,      
        colframe=black,     
        arc=5pt,            
        boxrule=1pt,        
        width=\linewidth,   
        center,             
        left=0mm,           
        right=0mm,          
        top=0mm,            
        bottom=0mm,         
        title={The Prompt for Informal-to-Formal Reasoning Synthesis}
    ]

You are given a natural language math problem along with its corresponding formal theorem statement in Lean 4 code. You need to provide a step-by-step analysis and reasoning of the process that bridges the gap from the natural language formulation to its Lean 4 formalization, including but not limited to the following parts:

\medskip

1. Start with a succinct restatement of the natural language math problem.

2. Outline the high-level logical structure of the mathematical problem, such as if-and-only-if, implication, universal quantification, or conjunction. Then, offer an in-depth analysis of the formalization process specific to this type of logical structure, carefully identifying any subtleties, potential pitfalls, or technical considerations that must be addressed in Lean 4.

3. Analyze the mathematical concepts involved in the problem using natural language.

4. Carefully analyze and then determine the mathematical objects to be defined and their corresponding types (e.g., natural numbers) based on the mathematical concepts.

5. Analyze the headers needed before formalizing the problem into a Lean 4 theorem, including importing libraries, opening namespaces, executing additional commands, or defining auxiliary functions, etc.

6. Anticipate tricky Lean issues (e.g., implicit type coercions, automatic variable declarations, operator overloading, etc.). 

7. Map the mathematical concepts, objects and operations mentioned in the problem to their corresponding representations in Lean 4. Before mapping each natural language mathematical item, first analyze its correspondence with the relevant Lean 4 objects in as much detail as possible.

8. Finally, review and output the complete Lean 4 code, ensuring it is fully consistent with the given formal theorem statement.

\medskip

Depth requirement:

1. Each item should be as detailed as possible, and each subpoint should contain more than three sentences.

2. Your reasoning must focus entirely on **formalization**, that is, how to translate the natural language problem into Lean 4 code step by step, and should NOT include any analysis of how to prove the statement.

Your response should be entirely in the tone of imitating the reasoning process, without exposing the task description, such as using terms like ``provided code'' or ``given code'' Avoid using such expressions.

\medskip

Here are some examples. Your reasoning should be more detailed and much longer than them ($\sim$ 4000 words):

\medskip

[Example 1]

\{Few\_Shots\_1\}

\medskip

[Example 2]

\{Few\_Shots\_2\}

\medskip

[Example 3]

\{Few\_Shots\_3\}

\medskip

\medskip

[YOUR TASK HERE]

\#\#\# Natural Language Problem

\{informal\_statement\}

\medskip

\#\#\# Lean 4 Code

\textasciigrave\textasciigrave\textasciigrave Lean4

\{formal\_statement\}

\textasciigrave\textasciigrave\textasciigrave

\end{tcolorbox}

\end{tcolorbox}
\caption{The complete prompt for informal-to-formal reasoning trajectory synthesis.}
\label{fig:prompt_reasoning_synthesis}
\end{figure*}
\onecolumn
    \begin{tcolorbox}[
        colback=white,      
        colframe=black,     
        arc=5pt,            
        boxrule=1pt,        
        width=\linewidth,   
        center,             
        left=0mm,           
        right=0mm,          
        top=0mm,            
        bottom=0mm,         
        title={Few-Shot Example 1 in Reasoning Trajectory Synthesis.},
        breakable
    ]

\#\#\# Natural Language Problem:

Suppose that $f$ is holomorphic in an open set $\Omega$. Prove that if $\text{Re}(f)$ is constant, then $f$ is constant.

\#\#\# Lean 4 Code:

\textasciigrave\textasciigrave\textasciigrave Lean4
\begin{lstlisting}[language=lean,basicstyle=\fontsize{8}{8}\ttfamily,frame=none,escapechar=!,breakindent=0pt]
import Mathlib

open Complex Function Metric
open scoped Topology

theorem my_favorite_theorem {f : ℂ → ℂ} (Ω : Set ℂ) (a b : Ω) (h : IsOpen Ω)
  (hf : DifferentiableOn ℂ f Ω) (hc : ∃ (c : ℝ), ∀ z ∈ Ω, (f z).re = c) :
  f a = f b :=
sorry
\end{lstlisting}
\textasciigrave\textasciigrave\textasciigrave

[Response]

We should formalize the following theorem:

If a function $f$ is holomorphic in an open set $\Omega$ and its real part $\text{Re}(f)$ is constant, then $f$ must be a constant function.

\medskip

The problem involves a conditional statement, which is a typical "if-then" structure. More formally, it follows an implication of two assumption leading to a conclusion:

Assumption: 

1. Function $f$ is holomorphic in an open set $\Omega$.

2. The real part of $f$ is constant.

Conclusion:

$f$ is constant on $\Omega$.

In addition, "$\text{Re}(f)$ is constant" implies that for any input, the real part of $f$ is the same. This needs to be described using both the existential quantifier ($\exists$) and the universal quantifier ($\forall$).

\medskip

There are some mathematical concepts in the problem: 

1. holomorphic

A holomorphic function is a complex-valued function of one or more complex variables that is complex differentiable in a neighbourhood of each point in a domain in complex coordinate space.

2. open set

An open set in a metric space is a set that, with every point P in it, contains all points of the metric space that are sufficiently near to P.

3. $\text{Re}$

It means taking the real part of a complex function.

4. constant

The first ``constant'' refers to a fixed number, a value that remains unchanged. The second ``constant'' refers to a constant function, which is a type of function that always produces the same output, regardless of the input.

\medskip

By analyzing the variable names in the problem, we find that it involves the following mathematical entities: 

1. f

Since the problem mentions that f is ``holomorphic'', it should be a mapping from complex numbers to complex numbers.

2. $\Omega$

A set of complex numbers.

Now we can do some preparatory work for formalizing this problem in Lean 4:

First, we need to import the mathematical library \textasciigrave Mathlib\textasciigrave. Since both f and $\Omega$ are defined on the complex domain, the \textasciigrave Complex\textasciigrave namespace is required. Additionally, since the problem involves functions and metric spaces, we also need the \textasciigrave Function\textasciigrave and \textasciigrave Metric\textasciigrave namespaces. Furthermore, since the open set is a concept in topology, we will need the \textasciigrave Topology\textasciigrave namespace as well. In summary, we need to execute the following Lean 4 code before formalizing the problem:
\textasciigrave\textasciigrave\textasciigrave Lean4

\begin{lstlisting}[language=lean,basicstyle=\fontsize{8}{8}\ttfamily,frame=none,escapechar=!,breakindent=0pt]
import Mathlib

open Complex Function Metric
open scoped Topology
\end{lstlisting}
\textasciigrave\textasciigrave\textasciigrave

\medskip

Next, There are several potential issues when formalizing this proof in Lean:

1. Implicit type coercion.

Lean's handling of complex numbers and real numbers can sometimes introduce implicit coercions. Specifically, $f$ is a complex-valued function, and we need to ensure that Lean correctly interprets the real part of $f$ as a real number.

2. Type class resolution.

We must ensure that Lean can resolve the necessary type classes, such as differentiability and the real part of a complex function.

3. Correctness of quantification.

We are quantifying over elements of the set $\Omega$, so we must ensure that the universal quantifier is applied properly to all complex number in $\Omega$.

\medskip

Then, let's analyze the Lean 4 code corresponding to the specific concepts and objects in the problem:

1. An open set $\Omega$ on the complex domain.
In Lean, a set in the complex domain is represented by \textasciigrave $\Omega$ : Set $\mathbb{C}$\textasciigrave, which means $\Omega$ is a collection of complex numbers. We use \textasciigrave IsOpen\textasciigrave predicate to ensure the topological open property for $\Omega$.

\textasciigrave\textasciigrave\textasciigrave
\begin{lstlisting}[language=lean,basicstyle=\fontsize{8}{8}\ttfamily,frame=none,escapechar=!,breakindent=0pt]
Ω : Set ℂ
h : IsOpen Ω
\end{lstlisting}
\textasciigrave\textasciigrave\textasciigrave

2. A holomorphic function f.

We first establish that f is defined in the complex plane, i.e. \textasciigrave $f : \mathbb{C} \rightarrow \mathbb{C}$\textasciigrave. Next, we focus on the holomorphic property. In Lean, the \textasciigrave DifferentiableOn\textasciigrave predicate checks that f is complex differentiable at every point in the open set $\Omega$, which ensures that f is holomorphic. 

\textasciigrave\textasciigrave\textasciigrave
\begin{lstlisting}[language=lean,basicstyle=\fontsize{8}{8}\ttfamily,frame=none,escapechar=!,breakindent=0pt]
f : ℂ → ℂ
hf : DifferentiableOn ℂ f Ω
\end{lstlisting}
\textasciigrave\textasciigrave\textasciigrave

3. $\text{Re}(f)$ is constant.
It means that there exists ($\exists$) a real number c such that for any ($\forall$) z in $\Omega$, the real part of f(z) is equal to c. We use this ``existential-universal'' logical structure to describe ``the real part of f is constant.''

\textasciigrave\textasciigrave\textasciigrave
\begin{lstlisting}[language=lean,basicstyle=\fontsize{8}{8}\ttfamily,frame=none,escapechar=!,breakindent=0pt]
hc : ∃ (c : ℝ), ∀ z ∈ Ω, (f z).re = c
\end{lstlisting}
\textasciigrave\textasciigrave\textasciigrave

4. $f$ is a constant function.
The final goal is to prove that the function f is constant on $\Omega$, meaning that f takes the same value at any two points a and b within $\Omega$. In other words, the function f satisfies the condition f(a) = f(b), which would imply that f is constant on $\Omega$. We can formalize the goal as \textasciigrave $\forall a\ b : \Omega, f\  a = f\  b$\textasciigrave, but alternatively, we place \textasciigrave $a\ b : \Omega$ \textasciigrave in the conditions, as this better aligns with common expression practices.

\textasciigrave\textasciigrave\textasciigrave
\begin{lstlisting}[language=lean,basicstyle=\fontsize{8}{8}\ttfamily,frame=none,escapechar=!,breakindent=0pt]
a b : Ω
f a = f b
\end{lstlisting}
\textasciigrave\textasciigrave\textasciigrave

\medskip

In conclusion, the complete Lean 4 code is as follows:

\textasciigrave\textasciigrave\textasciigrave Lean4
\begin{lstlisting}[language=lean,basicstyle=\fontsize{8}{8}\ttfamily,frame=none,escapechar=!,breakindent=0pt]
import Mathlib

open Complex Function Metric
open scoped Topology

theorem my_favorite_theorem {f : ℂ → ℂ} (Ω : Set ℂ) (a b : Ω) (h : IsOpen Ω)
  (hf : DifferentiableOn ℂ f Ω) (hc : ∃ (c : ℝ), ∀ z ∈ Ω, (f z).re = c) :
  f a = f b :=
sorry
\end{lstlisting}
\textasciigrave\textasciigrave\textasciigrave

\end{tcolorbox}
\captionof{figure}{The first few-shot example for informal-to-formal reasoning trajectory synthesis.}
\label{fig:prompt_reasoning_synthesis_few_shot_1}
\twocolumn 
\onecolumn
    \begin{tcolorbox}[
        colback=white,      
        colframe=black,     
        arc=5pt,            
        boxrule=1pt,        
        width=\linewidth,   
        center,             
        left=0mm,           
        right=0mm,          
        top=0mm,            
        bottom=0mm,         
        title={Few-Shot Example 2 in Reasoning Trajectory Synthesis.},
        breakable
    ]

\#\#\# Natural Language Problem:

If $\mathcal{T}_\alpha$ is a family of topologies on $X$, show that $\bigcap \mathcal{T}_\alpha$ is a topology on $X$.

\medskip

\#\#\# Lean 4 Code:

\GR\GR\GR Lean4
\begin{lstlisting}[language=lean,basicstyle=\fontsize{8}{8}\ttfamily,frame=none,escapechar=!,breakindent=0pt]
import Mathlib

open Set TopologicalSpace
open scoped Topology

def is_topology (X : Type*) (T : Set (Set X)) :=
  ∅ ∈ T ∧ univ ∈ T ∧
  (∀ s t, s ∈ T → t ∈ T → s ∩ t ∈ T) ∧
  (∀s, (∀t ∈ s, t ∈ T) → sUnion s ∈ T)

theorem my_favorite_theorem (X I : Type*) (T : I → Set (Set X)) (h : ∀ i, is_topology X (T i)) :
  is_topology X (⋂ i : I, T i) :=
sorry
\end{lstlisting}
\GR\GR\GR

\medskip

[Response]

Let's restate the natural language math problem:

The intersection of a family of topologies on $X$, i.e. $\bigcap \mathcal{T}_\alpha$, is also a topology on $X$.

\medskip

The logical structure of the problem involves a universal quantification and an implication:

1. Universal quantification ($\forall$).

For each index $\alpha$, $\mathcal{T}_\alpha$ is a topology on $X$.

2. Implication.

The problem follows an ``if-then'' structure, where the assumption is that $\mathcal{T}_\alpha$ is a family of topologies, and the conclusion is that the intersection of these topologies is also a topology. We should formalize these two parts separately.

\medskip

The problem involves two mathematical concepts, topology and $\bigcap$:

1. Topologies on sets.

Let X be a set and let T be a family of subsets of X. Then T is called a topology on X if:

(1) Both the empty set and X are elements of T.

(2) Any intersection of finitely many elements of T is an element of T.

(3) Any union of elements of T is an element of T.

2. $\bigcap \mathcal{T}_\alpha$

It means the intersection of all sets in T.

\medskip

We also identify these mathematical entities in the natural language problem:

1. Set $X$ and its set family.

We use \GR set X\GR~to denote a set of elements of an arbitrary type, and \GR set (set X)\GR~to denote the set of all its subsets, i.e., a family of sets.

2. A topology on $X$.

According to the definition of a topology, its type is also \GR set (set X)\GR.

3. A family of topologies.

It means a set of all topologies on $X$. We can use a mapping from an index type $I$ to a family of sets to define a family of topologies.

\medskip

Before considering the formalization of the statement, we should add some helper code.

We should import \GR Mathlib\GR~first. Since the problem is about the topologies on set, we can open namespaces \GR Set\GR, \GR TopologicalSpace\GR~ and \GR Topology\GR. When we are already in these namespaces, we can define an auxiliary function \GR is\_topology\GR to simplify the formalization of the problem. The input of this function is a family of sets \GR T\GR, and the output is whether it satisfies the three defining properties of a topology.

Here are the helper code:

\GR\GR\GR Lean4
\begin{lstlisting}[language=lean,basicstyle=\fontsize{8}{8}\ttfamily,frame=none,escapechar=!,breakindent=0pt]
import Mathlib

open Set TopologicalSpace
open scoped Topology

def is_topology (X : Type*) (T : Set (Set X)) :=
  ∅ ∈ T ∧ univ ∈ T ∧
  (∀ s t, s ∈ T → t ∈ T → s ∩ t ∈ T) ∧
  (∀s, (∀t ∈ s, t ∈ T) → sUnion s ∈ T)
\end{lstlisting}
\GR\GR\GR

\medskip

We should pay attention to the following potential formalization issues in Lean 4:

1. Function application in dependent types.
Since the family of topologies should be formalized as a function like \GR T : I $\rightarrow$ Set (Set X)\GR~in Lean, we must be careful with how we apply it and refer to elements of \GR T i\GR~. Mistakenly writing or misplacing parentheses could lead to type mismatch errors. Lean is strict about such syntax.

2. Make full use of already defined functions.
So far, we have defined the auxiliary function \GR is\_topology\GR~ to determine whether a family of sets forms a topology. We should invoke it in both the conditions and the conclusion of the formalization, rather than repeatedly write out the definition of a topology.

\medskip

Next, let's examine the different parts of Lean 4 code that corresponds to the problem:

1. Declaration of $\mathcal{T}_\alpha$.

We need a type \GR X\GR~to represent the underlying set over which the topologies are defined. Additionally, we introduce a type \GR I\GR~to serve as the indexing set for the family of topologies, which corresponds to the symbol $\alpha$ in the natural language formulation. Building on these type declarations, we can use the function \GR T : I $\rightarrow$ Set (Set X)\GR~to represent the family of topologies.

\GR\GR\GR
\begin{lstlisting}[language=lean,basicstyle=\fontsize{8}{8}\ttfamily,frame=none,escapechar=!,breakindent=0pt]
X I : Type*
T : I → Set (Set X)
\end{lstlisting}
\GR\GR\GR

2. $\mathcal{T}_\alpha$ is a family of topologies on $X$.

It is the premise of the proposition to be formalized. We consider over all indices \GR i\GR~and apply \GR is\_topology\GR~to ensure that each corresponding collection \GR T i\GR~is a topology under type \GR X\GR.

\GR\GR\GR
\begin{lstlisting}[language=lean,basicstyle=\fontsize{8}{8}\ttfamily,frame=none,escapechar=!,breakindent=0pt]
h : ∀ i, is_topology X (T i)
\end{lstlisting}
\GR\GR\GR

3. $\bigcap \mathcal{T}_\alpha$ is a topology on $X$.

It is the conclusion. We use \GR($\bigcap$ i : I, T i)\GR~to represent the intersection of all the topologies, and similarly apply \GR is\_topology\GR~ to claim that it is also a topology.

\GR\GR\GR
\begin{lstlisting}[language=lean,basicstyle=\fontsize{8}{8}\ttfamily,frame=none,escapechar=!,breakindent=0pt]
is_topology X (⋂ i : I, T i)
\end{lstlisting}
\GR\GR\GR

\medskip

Above all, the complete Lean 4 code is as follows:

\GR\GR\GR Lean4
\begin{lstlisting}[language=lean,basicstyle=\fontsize{8}{8}\ttfamily,frame=none,escapechar=!,breakindent=0pt]
import Mathlib

open Set TopologicalSpace
open scoped Topology

def is_topology (X : Type*) (T : Set (Set X)) :=
  ∅ ∈ T ∧ univ ∈ T ∧
  (∀ s t, s ∈ T → t ∈ T → s ∩ t ∈ T) ∧
  (∀s, (∀t ∈ s, t ∈ T) → sUnion s ∈ T)

theorem my_favorite_theorem (X I : Type*) (T : I → Set (Set X)) (h : ∀ i, is_topology X (T i)) :
  is_topology X (⋂ i : I, T i) :=
sorry
\end{lstlisting}
\GR\GR\GR

\end{tcolorbox}
\captionof{figure}{The second few-shot example for informal-to-formal reasoning trajectory synthesis.}
\label{fig:prompt_reasoning_synthesis_few_shot_2}
\twocolumn 
\onecolumn
    \begin{tcolorbox}[
        colback=white,      
        colframe=black,     
        arc=5pt,            
        boxrule=1pt,        
        width=\linewidth,   
        center,             
        left=0mm,           
        right=0mm,          
        top=0mm,            
        bottom=0mm,         
        title={Few-Shot Example 3 in Reasoning Trajectory Synthesis.},
        breakable
    ]

\#\#\# Natural Language Problem:

Prove that the Cauchy product of two absolutely convergent real series converges absolutely.

\medskip

\#\#\# Lean 4 Code:

\GR\GR\GR Lean4
\begin{lstlisting}[language=lean,basicstyle=\fontsize{8}{8}\ttfamily,frame=none,escapechar=!,breakindent=0pt]
import Mathlib

open Topology Filter Real Finset
open scoped BigOperators

theorem my_favorite_theorem
  (a b : ℕ → ℝ)
  (ha : ∃ y, (Tendsto (λ n => (∑ i in (range n), |a i|)) atTop (__N y)))
  (hb : ∃ y, (Tendsto (λ n => (∑ i in (range n), |b i|)) atTop (__N y))) :
  ∃ y, (Tendsto (λ n => (∑ i in (range n),
  λ i => (∑ j in range (i + 1), |a j * b (i - j)|))) atTop (__N y)) :=
sorry
\end{lstlisting}
\GR\GR\GR

\medskip

[Response]

We are asked to formalize the following natural language statement to Lean 4:

If two infinite series made from sequences of real numbers are absolutely convergent, then the series formed by multiplying them term by term in the Cauchy product sense also converges absolutely.

\medskip

The logical structure of the problem is a classic implication involving existential quantification ($\exists$) and series convergence:

1. Premises: Two infinite series are absolutely convergent, meaning the limit of the sum of the absolute values of their terms exists.

2. Conclusion: The Cauchy product of the two series converges absolutely.

\medskip

In this problem, we encounter some mathematical concepts:

1. Cauchy product

Let $\sum_{i=0}^{\infty} a_{i}$ and $\sum_{j=0}^{\infty} b_{j}$ be two infinite series with real terms. The Cauchy product of these two infinite series is defined by a discrete convolution as follows:
$$
	\left(\sum_{i=0}^{\infty} a_{i}\right) \cdot \left(\sum_{j=0}^{\infty} b_{j}\right) = \sum_{k=0}^{\infty} c_{k}
$$

where $c_{k} = \sum_{l=0}^{k} a_{l}b_{k-l}$.

2. Absolutely convergent series

An infinite series of numbers $a_{n}$ is said to converge absolutely (or to be absolutely convergent) if $\sum_{n=0}^{\infty} \left| a_{n} \right| = L$ for some real number $L$.

\medskip

Through analyzing the variables in the problem, we recognize that it involves these mathematical entities:

1. Two real series

We represent these two series using mappings from natural numbers to real numbers, and denote them as a and b.

2. The Cauchy product

Directly compute the Cauchy product of a and b using discrete convolution. This is also a mapping, and for simplicity, we can represent it using a $\lambda$-expression.

3. The limit of the partial sums of a series.

We need to characterize the absolute convergence of a series by the existence of the limit of its partial sums. This limit value must be explicitly defined using an existential quantifier.

\medskip

Prior to formalizing the statement, we need to include some auxiliary code.

With Mathlib imported, we characterize the convergence of a series by considering the limit as they tend to infinity. This can be done using \GR Tendsto\GR~and \GR atTop\GR~from the \GR Filter\GR~module, so we need to open the \GR Filter\GR~namespace. We also need \GR Topology\GR~to use $\mathcal{N}$ to describe the neighborhood. Since the problem is considered in the real number domain, we need to open \GR Real\GR~namespace. For convenience in expression, we also use \GR Finset\GR~and \GR BigOperators\GR~to represent summation over a specific range.

The Lean 4 code is shown below:

\GR\GR\GR lean4
\begin{lstlisting}[language=lean,basicstyle=\fontsize{8}{8}\ttfamily,frame=none,escapechar=!,breakindent=0pt]
import Mathlib

open Filter Topology Real Finset
open scoped BigOperators
\end{lstlisting}
\GR\GR\GR

\medskip

Before delving into the details of Lean 4, we need to be mindful of the following points:

1. Managing nested sums and indexing dependencies.

The Cauchy product involves double summation where inner indices depend on outer ones. It's crucial to ensure that index bounds are valid to prevent undefined expressions.

2. Formalizing convergence with filters.

Lean uses \GR Tendsto\GR~and the \GR atTop\GR~filter to express convergence, which refers to the convergence of partial sums. Confusing this with convergence of the series itself can lead to incorrect formulations.

3. Balancing implicit inference and explicit types.

Lean performs implicit type inference for notations like \GR$\sum$\GR~and \GR$|\cdot|$\GR~, but this can fail in complex expressions. Adding explicit type annotations when needed helps prevent type-checking errors.

\medskip

Now, let's break down the Lean 4 code associated with the specific concepts and objects in the problem:

1. Two real series.

In Lean 4, we represent the two infinite series \GR a\GR~and \GR b\GR~as functions from the set of natural numbers \GR$\mathbb{N}$\GR~to the set of real numbers \GR$\mathbb{R}$\GR~.

\GR\GR\GR
\begin{lstlisting}[language=lean,basicstyle=\fontsize{8}{8}\ttfamily,frame=none,escapechar=!,breakindent=0pt]
a b : ℕ → ℝ
\end{lstlisting}
\GR\GR\GR

2. The counterparts of the partial sums of the absolute series. 
For a fixed n, the expression \GR $\sum$ i in range n, $|$ a i $|$\GR~computes $\sum_{i=0}^{n-1} \left| a_{i} \right|$, which is how the partial sums are formalized. We formalize them in terms of $\lambda$-expression for the further formalization of absolute convergence.

\GR\GR\GR
\begin{lstlisting}[language=lean,basicstyle=\fontsize{8}{8}\ttfamily,frame=none,escapechar=!,breakindent=0pt]
λ n => (∑ i in (range n), |a i|
λ n => (∑ i in (range n), |b i|
\end{lstlisting}
\GR\GR\GR

3. Absolutely convergent series.

We should formalize that $\sum_{i=0}^{\infty} \left| a_{i} \right|$ converges. In Lean, this is expressed by saying that the sequence of partial sums tends to some real limit \GR y\GR~as \GR n\GR~tends to infinity, using the \GR Tendsto\GR~predicate, the \GR atTop\GR~filter, and the \GR$\mathcal{N}$ y\GR~notation:

In Lean, \GR Tendsto\GR~is a predicate used to formalize the notion of convergence of functions with respect to filters. Specifically, \GR Tendsto f $l_1$ $l_2$\GR~means that the function \GR f\GR~tends to the filter \GR$l_2$` as its input varies along the filter \GR$l_1$\GR.

The \GR atTop\GR~filter on natural numbers captures the idea of ``tending towards infinity''. In other words, \GR atTop\GR~describes the behavior of a sequence as its index grows without bound.

The notation \GR $\mathcal{N}$ y\GR~denotes the neighborhood filter of the point \GR y\GR~in a topological space (here, the real numbers $\mathbb{R}$).

\GR\GR\GR
\begin{lstlisting}[language=lean,basicstyle=\fontsize{8}{8}\ttfamily,frame=none,escapechar=!,breakindent=0pt]
ha : ∃ y, (Tendsto (λ n => (∑ i in (range n), |a i|)) atTop (__N y))
hb : ∃ y, (Tendsto (λ n => (∑ i in (range n), |b i|)) atTop (__N y))
\end{lstlisting}
\GR\GR\GR

4. Their Cauchy product.

For each n, the outer sum computes $\sum_{i=0}^{n-1} c_i$, where each $c_i$ is itself a sum of products $\sum_{j=0}^{i} \left| a_{j} b_{i-j}\right|$; that inner sum is exactly the j-indexed expression inside. Be mindful of indexing: since \GR Finset.range k\GR represents the half-open interval [0, k), if we want to include k in the range, we must use \GR Finset.range (k + 1)\GR instead.

\GR\GR\GR
\begin{lstlisting}[language=lean,basicstyle=\fontsize{8}{8}\ttfamily,frame=none,escapechar=!,breakindent=0pt]
λ n => (∑ i in (range n),
  λ i => (∑ j in range (i + 1), |a j * b (i - j)|))
\end{lstlisting}
\GR\GR\GR

5. The absolute convergence of the Cauchy product.

Similar to the earlier formalization of absolute convergence.

\GR\GR\GR
\begin{lstlisting}[language=lean,basicstyle=\fontsize{8}{8}\ttfamily,frame=none,escapechar=!,breakindent=0pt]
∃ y, (Tendsto (λ n => (∑ i in (range n),
  λ i => (∑ j in range (i + 1), |a j * b (i - j)|))) atTop (__N y))
\end{lstlisting}
\GR\GR\GR

In summary, the complete Lean 4 code is outlined as follows: 

\GR\GR\GR Lean4
\begin{lstlisting}[language=lean,basicstyle=\fontsize{8}{8}\ttfamily,frame=none,escapechar=!,breakindent=0pt]
import Mathlib

open Topology Filter Real Finset
open scoped BigOperators

theorem my_favorite_theorem
  (a b : ℕ → ℝ)
  (ha : ∃ y, (Tendsto (λ n => (∑ i in (range n), |a i|)) atTop (__N y)))
  (hb : ∃ y, (Tendsto (λ n => (∑ i in (range n), |b i|)) atTop (__N y))) :
  ∃ y, (Tendsto (λ n => (∑ i in (range n),
  λ i => (∑ j in range (i + 1), |a j * b (i - j)|))) atTop (__N y)) :=
sorry
\end{lstlisting}
\GR\GR\GR

\end{tcolorbox}
\captionof{figure}{The third few-shot example for informal-to-formal reasoning trajectory synthesis.}
\label{fig:prompt_reasoning_synthesis_few_shot_3}
\twocolumn 
\begin{figure*}[t]
    \centering    
    \begin{tcolorbox}[
        colback=white,      
        colframe=white,     
        arc=0pt,            
        boxrule=1pt,        
        width=\linewidth,
        center,             
        left=0mm,           
        right=0mm,          
        top=0mm,            
        bottom=0mm,         
    ]
    \begin{tcolorbox}[
        colback=white,      
        colframe=black,     
        arc=5pt,            
        boxrule=1pt,        
        width=\linewidth,   
        center,             
        left=0mm,           
        right=0mm,          
        top=0mm,            
        bottom=0mm,         
        title={Off-Task Example of Claude4-thinking}
    ]

    \textbf{Prompt}: Please autoformalize the following problem in Lean 4 with a header. Use the following theorem names: my\_favorite\_theorem. 

\medskip  
Let \(a\) and \(b\) be positive integers and satisfy the relation
\[
(11111 + a)(11111 - b)=123456789.
\]
Prove that \(a - b\) is a multiple of \(4\).

\medskip
Your code should start with:

\textasciigrave\textasciigrave\textasciigrave Lean4

import Mathlib

\textasciigrave\textasciigrave\textasciigrave

\medskip

You should only output the theorem statement in Lean 4 format, ending with \textasciigrave sorry \textasciigrave. You should NOT output the proof.

    \medskip
    \selectfont
    \textbf{Response} (Claude4-thinking):
    
    \textless think\textgreater
    
\fontsize{8}{10}

Let me think about this problem. We have:

- $a$ and $b$ are positive integers

- $(11111 + a)(11111 - b) = 123456789$

We need to prove that $a - b$ is a multiple of 4.

Let me first expand the left side:

$(11111 + a)(11111 - b) = 11111^2 + a \cdot 11111 - b \cdot 11111 - ab$

$= 11111^2 + 11111(a - b) - ab$

So we have:

$11111^2 + 11111(a - b) - ab = 123456789$

Let me calculate $11111^2$:

$11111^2 = 123454321$

So:

$123454321 + 11111(a - b) - ab = 123456789$

$11111(a - b) - ab = 123456789 - 123454321$

$11111(a - b) - ab = 2468$

So we have:

$11111(a - b) - ab = 2468$

Now let me think about this modulo 4. 

First, what is $11111 \pmod 4$?

$11111 = 4 \cdot 2777 + 3$, so $11111 \equiv 3 \pmod 4$.

What is $2468 \pmod 4$?

$2468 = 4 \cdot 617$, so $2468 \equiv 0 \pmod 4$.

So we have:

$11111(a - b) - ab \equiv 0 \pmod 4$

$3(a - b) - ab \equiv 0 \pmod 4$

Now let me think about this differently. Let's consider $a$ and $b$ modulo 4.

...

\textless /think\textgreater

Looking at this problem, I need to formalize the statement that for positive integers $a$ and $b$ satisfying $(11111 + a)(11111 - b) = 123456789$, we have that $a - b$ is divisible by 4.

\textasciigrave\textasciigrave\textasciigrave Lean4

    \begin{lstlisting}[language=lean,basicstyle=\fontsize{8}{8}\ttfamily,frame=none,escapechar=!,breakindent=0pt]
import Mathlib

theorem my_favorite_theorem (a b : ℤ) 
  (ha : 0 < a) (hb : 0 < b)
  (h : (11111 + a) * (11111 - b) = 123456789) : 
  4 ∣ (a - b) := by
  sorry
\end{lstlisting}
    
\textasciigrave\textasciigrave\textasciigrave

    \end{tcolorbox}

\end{tcolorbox}
\caption{\textbf{An example of off-task during formalization (Claude4-thinking).} It spends an extremely long reasoning process on solving the natural language problem itself rather than translating it into formal language. Strangely, it still outputs the correct formal statement in the end.}
\label{fig:claude4_bad_example}
\end{figure*}

\onecolumn
    \begin{tcolorbox}[
        colback=white,      
        colframe=black,     
        arc=5pt,            
        boxrule=1pt,        
        width=\linewidth,   
        center,             
        left=0mm,           
        right=0mm,          
        top=0mm,            
        bottom=0mm,         
        title={Case Study: abstract\_algebra\_\_p15 (ProverBench)},
        breakable
    ]

\textbf{Informal Problem:}

Let $\mathbb{Z}_3$ be the finite field of integers modulo 3, and let $c \in \mathbb{Z}_3$. The quotient ring $\mathbb{Z}_3[x]/(x^3 + x^2 + c)$ is a field if and only if $c = 2$.

\medskip

\textbf{Headers:}

\textasciigrave\textasciigrave\textasciigrave Lean4
\begin{lstlisting}[language=lean,basicstyle=\fontsize{8}{8}\ttfamily,frame=none,escapechar=!,breakindent=0pt]
import Mathlib
open Polynomial Ideal

variable {R : Type*} [CommRing R]
\end{lstlisting}
\textasciigrave\textasciigrave\textasciigrave

\medskip
\textbf{\KMAuto-7B:}

\textasciigrave\textasciigrave\textasciigrave Lean4
\begin{lstlisting}[language=lean,basicstyle=\fontsize{8}{8}\ttfamily,frame=none,escapechar=!,breakindent=0pt]
import Mathlib
open Polynomial Ideal

variable {R : Type*} [CommRing R]
def ZMod3 : Type* := {x : R | x ^ 3 - 1 = 0}
def ZMod3_fintype : Fintype ZMod3 := {x : R | x ^ 3 - 1 = 0}
def ZMod3_quotient (c : ZMod3) : Type* := {x : ZMod3 | x ^ 3 + x ^ 2 + c = 0}
def ZMod3_quotient_fintype (c : ZMod3) : Fintype (ZMod3_quotient c) := {x : ZMod3 | x ^ 3 + x ^ 2 + c = 0}
theorem my_favorite_theorem (c : ZMod3) :
    IsField (ZMod3_quotient c) ↔ c = 2 := by sorry
\end{lstlisting}
\textasciigrave\textasciigrave\textasciigrave

(Syntax Error)

\medskip

\textbf{\modelname-7B (Ours):}

\textless think\textgreater

\# Formalizing the Field Property of Quotient Rings in Lean 4

Let me restate the natural language problem succinctly:

The quotient ring $\mathbb{Z}_3[x]/(x^3 + x^2 + c)$ is a field if and only if $c = 2$ (where $c \in \mathbb{Z}_3$).

\medskip

\#\# Logical Structure Analysis

This problem follows an if-and-only-if (biconditional) logical structure, which is represented by the ``$\leftrightarrow$'' symbol in Lean. The biconditional statement connects two propositions:

1. Left side: The quotient ring $\mathbb{Z}_3[x]/(x^3 + x^2 + c)$ is a field.

2. Right side: $c = 2$.

Formalizing a biconditional statement in Lean 4 requires proving both directions of the implication:

- Forward direction: If the quotient ring is a field, then $c = 2$.

- Backward direction: If $c = 2$, then the quotient ring is a field.

This logical structure is particularly important in algebraic statements where we need to establish equivalence between algebraic properties and specific parameter values. The biconditional ensures that we capture exactly when the algebraic structure has the desired property, without including unnecessary cases or excluding valid ones.

\medskip

\#\# Mathematical Concepts Analysis

The problem involves several mathematical concepts:

1. Finite field $\mathbb{Z}_3$: This is the field of integers modulo 3, consisting of the elements {0, 1, 2} with arithmetic operations performed modulo 3.

2. Polynomial ring $\mathbb{Z}_3[x]$: This is the ring of polynomials with coefficients from $\mathbb{Z}_3$. Elements have the form $a_n x^n + a_{n-1} x^{n-1} + \ldots + a_1 x + a_0$ where each $a_i \in \mathbb{Z}_3$.

3. Ideal generated by a polynomial: The ideal $(x^3 + x^2 + c)$ consists of all polynomials that are multiples of $x^3 + x^2 + c$.

4. Quotient ring: The quotient ring $\mathbb{Z}_3[x]/(x^3 + x^2 + c)$ consists of equivalence classes of polynomials where two polynomials are equivalent if their difference is in the ideal $(x^3 + x^2 + c)$.

5. Field: A field is a ring where every non-zero element has a multiplicative inverse. For a quotient ring $\mathbb{Z}_3[x]/(p(x))$ to be a field, the polynomial $p(x)$ must be irreducible over $\mathbb{Z}_3$.

\medskip

\#\# Mathematical Objects and Types

Based on the problem, we need to define or work with the following mathematical objects:

1. The ring $\mathbb{Z}_3$: In Lean, this is represented by the type \GR ZMod 3\GR, which is the ring of integers modulo 3.

2. Polynomials over $\mathbb{Z}_3$: These are represented by the type \textasciigrave Polynomial (ZMod 3)\textasciigrave in Lean.

3. The specific polynomial $x^3 + x^2 + c$: This is constructed using polynomial operations in Lean.

4. The ideal generated by this polynomial: This is represented using Lean's ideal notation.

5. The quotient ring: This is represented using Lean's quotient ring construction.

\medskip

\#\# Required Headers and Imports

To formalize this problem in Lean 4, we need to import appropriate libraries and open relevant namespaces:

1. Mathlib: This is the standard mathematical library for Lean, which provides definitions and theorems about rings, fields, polynomials, etc.

2. Polynomial namespace: This provides operations and properties of polynomials.

3. Ideal namespace: This provides operations and properties of ideals in rings.

4. CommRing namespace: This provides properties of commutative rings, which are relevant since we're working with fields and quotient rings.

The import statement and namespace openings would look like:

\textasciigrave\textasciigrave\textasciigrave
\begin{lstlisting}[language=lean,basicstyle=\fontsize{8}{8}\ttfamily,frame=none,escapechar=!,breakindent=0pt]
import Mathlib
open Polynomial Ideal
\end{lstlisting}
\textasciigrave\textasciigrave\textasciigrave

\medskip

\#\# Potential Lean Issues

When formalizing this problem in Lean 4, we need to be aware of several potential issues:

1. Type coercions: Lean might need to coerce between different types, such as between elements of \GR ZMod 3\GR and polynomials over \GR ZMod 3\GR.

2. Variable declarations: We need to properly declare the variable \GR c\GR~ and ensure it's interpreted as an element of \GR ZMod 3\GR.

3. Notation for polynomials: Lean has specific syntax for constructing polynomials, and we need to ensure we're using it correctly.

4. Quotient ring construction: Lean has specific syntax for constructing quotient rings, and we need to ensure we're using it correctly.

5. Field property: We need to use the appropriate Lean predicate to express that a ring is a field.

\medskip

\#\# Mapping Mathematical Concepts to Lean 4

Now, let's map the mathematical concepts from the problem to their Lean 4 representations:

1. The ring $\mathbb{Z}_3$:

In Lean, the ring of integers modulo 3 is represented by the type \GR \texttt{ZMod 3}\GR. This is a finite field with elements \{0, 1, 2\} and arithmetic operations performed modulo 3.

2. The polynomial $x^3 + x^2 + c$:

In Lean, polynomials are constructed using the \GR X\GR variable for the indeterminate and standard arithmetic operations. The polynomial $x^3 + x^2 + c$ would be represented as \GR\texttt{X\textasciicircum3 + X\textasciicircum2 + C c}\GR, where \GR\texttt{C c}\GR~ is the constant term lifted from the coefficient ring.

3. The ideal $(x^3 + x^2 + c)$:

In Lean, ideals are constructed using the ideal notation. The ideal generated by the polynomial $x^3 + x^2 + c$ would be represented as \GR\texttt{span (\{X\textasciicircum3 + X\textasciicircum2 + C c\}: Set (Polynomial (ZMod 3)))}\GR, where \GR span\GR is the function that constructs the ideal generated by a set of elements.

4. The quotient ring $\mathbb{Z}_3[x]/(x^3 + x^2 + c)$:
In Lean, quotient rings are constructed using the quotient notation. The quotient ring $\mathbb{Z}_3[x]/(x^3 + x^2 + c)$ would be represented as \GR \texttt{(Polynomial (ZMod 3)) / (span (\{X\textasciicircum3 + X\textasciicircum2 + C c\} : Set (Polynomial (ZMod 3))))}\GR.

5. The field property:
In Lean, the property of being a field is represented by the predicate \GR IsField\GR. So, the statement that the quotient ring is a field would be represented as \GR \texttt{IsField ((Polynomial (ZMod 3)) / (span (\{X\textasciicircum3 + X\textasciicircum2 + C c\} : Set (Polynomial (ZMod 3))))}\GR.

6. The condition $c = 2$:
   In Lean, this is simply represented as \GR\texttt{c = 2}\GR, where \GR c\GR is a variable of type \GR\texttt{ZMod 3}\GR.

\medskip

\#\# Complete Lean 4 Code

Based on the above analysis, the complete Lean 4 code for formalizing the problem would be:

\GR\GR\GR lean
\begin{lstlisting}[language=lean,basicstyle=\fontsize{8}{8}\ttfamily,frame=none,escapechar=!,breakindent=0pt]
import Mathlib
open Polynomial Ideal

variable {R : Type*} [CommRing R]
theorem my_favorite_theorem {c : ZMod 3} :
    IsField ((Polynomial (ZMod 3)) ⧸ (span ({X^3 + X^2 + C c} : Set (Polynomial (ZMod 3)))))
    ↔ c = 2 := by
  sorry
\end{lstlisting}
\GR\GR\GR

...

\textless /think\textgreater

\GR\GR\GR Lean4
\begin{lstlisting}[language=lean,basicstyle=\fontsize{8}{8}\ttfamily,frame=none,escapechar=!,breakindent=0pt]
import Mathlib
open Polynomial Ideal

variable {R : Type*} [CommRing R]
theorem my_favorite_theorem {c : ZMod 3} :
    IsField ((Polynomial (ZMod 3)) ⧸ (span ({X^3 + X^2 + C c} : Set (Polynomial (ZMod 3)))))
    ↔ c = 2 := by
  sorry
\end{lstlisting}
\GR\GR\GR

\end{tcolorbox}
\captionof{figure}{A comprehensive case study in the OOD benchmark ProverBench. The output of \KMAuto indicates that it could not understand the concept of a quotient ring, while \modelname correctly interprets the problem by analyzing the natural langauge definitions of polynomial rings and ideals, and successfully maps it to the corresponding Lean 4 constructs.}
\label{fig:case_study_1}
\twocolumn 

\end{document}